\documentclass{article} 
\usepackage[final]{colm2025_conference}

\usepackage[table,xcdraw]{xcolor}
\usepackage{microtype}
\usepackage{graphicx}
\usepackage{hyperref}
\usepackage{url}
\usepackage{multirow}
\usepackage{booktabs}
\usepackage{subcaption}

\usepackage[ruled,lined,linesnumbered]{algorithm2e}
\usepackage[noend]{algorithmic}

\usepackage{amsmath}
\usepackage{amssymb}
\usepackage{mathtools}
\usepackage{amsthm}

\usepackage{amsmath,amsfonts,bm}

\def\eqref#1{equation~\ref{#1}}

\def\1{\bm{1}}
\newcommand{\train}{\mathcal{D}}

\newcommand{\test}{\mathcal{D_{\mathrm{test}}}}

\def\ve{{\bm{e}}}

\DeclareMathAlphabet{\mathsfit}{\encodingdefault}{\sfdefault}{m}{sl}
\SetMathAlphabet{\mathsfit}{bold}{\encodingdefault}{\sfdefault}{bx}{n}

\usepackage{enumitem}
\usepackage{url}
\usepackage{fdsymbol}
\usepackage{cleveref}

\usepackage{lineno}

\definecolor{darkblue}{rgb}{0, 0, 0.5}
\hypersetup{colorlinks=true, citecolor=darkblue, linkcolor=darkblue, urlcolor=darkblue}

\title{RARe: Retrieval-Augmented Retrieval With \\ In-Context Examples}

\author{Atula Tejaswi$^{\spadesuit}$, Yoonsang Lee$^{\heartsuit}$, Sujay Sanghavi$^{\spadesuit*}$, Eunsol Choi$^{\diamondsuit}$\thanks{Equal advising.} \\
\small{$^{\spadesuit}$The University of Texas at Austin  
\ $^{\heartsuit}$Princeton University \ $^{\diamondsuit}$New York University }\\
\texttt{\url{atutej@utexas.edu}}\\
}

\begin{document}

\ifcolmsubmission
\linenumbers
\fi

\maketitle

\begin{abstract}
While in-context learning is well-studied with decoder-only language models (LLMs), its utility for encoder-only models remains underexplored. We study in-context learning for encoder-only models for text retrieval tasks. Can incorporating in-context examples (query-document pairs) to the target query enhance retriever performance? Our approach, \texttt{RARe}, finetunes a pre-trained model with in-context examples whose query is semantically similar to the target query. This approach achieves performance gains of up to +2.72\% nDCG across open-domain retrieval datasets (BeIR, RAR-b) compared to using the target query only as an input. In particular, we find \texttt{RARe} exhibits stronger out-of-domain generalization compared to models using queries without in-context examples, similar to what is seen for in-context learning in LLMs. We further provide analysis on the design choices of in-context example augmentation for retrievers and lay the foundation for future work. 
\end{abstract}

\section{Introduction}
In-context learning (ICL) \citep{brown2020language} has emerged as a powerful paradigm enabling diverse applications without parameter updates in large language models (LLMs). 
While in-context learning has been extensively studied for the decoder-only models \citep{xu2023knnpromptingbeyondcontextlearning, min-etal-2022-metaicl, dong2024surveyincontextlearning}, its application to encoder-only, embedding models have been limited. Recently, \citet{li2024makingtextembeddersfewshot} proposed to incorporate randomly sampled few-shot examples into the query side to enhance the query embedding for embedding models. They reported strong performances, achieving state-of-the-art performances in a popular retrieval benchmark~\citep{muennighoff-etal-2023-mteb}. In this paper, we further study how in-context examples enhance the performance in retriever models, exploring its design choices and generalization capability with controlled studies.

Compared to decoder models which generates new tokens from the input as a continuation, how in-context examples would impact the output vector from an embedding model is less obvious. Unlike in decoder-only LLMs where in-context examples expand model capacity at generation time, in-context examples in encoder-only models may primarily provide task-relevant information rather than increasing model capacity. We study injecting \textit{semantically similar} in-context examples to build a dense retriever model~\citep{karpukhin-etal-2020-dense} which embeds queries and documents into a shared representational space for efficient search over a large corpus. 
State-of-the-art retriever models started to leverage decoder-only models as a backbone \citep{wang2024improving, behnamghader2024llm2vec, muennighoff2024gritlm, SFRAIResearch2024, lee2024nvembedimprovedtechniquestraining}, further motivating our study of applying in-context examples. 

We begin by naively prepending in-context examples to the target query and provide it to existing retriever models~\citep{behnamghader2024llm2vec, wang2024improving, SFRAIResearch2024}. Unlike in decoder-only models, zero-shot modification leads to significant performance drop. We propose to construct retrieval models that can effectively leverage in-context examples, which we name as \texttt{RARe}: Retrieval Augmented Retrieval with In-Context Examples. Our approach modifies the query format of retrieval systems by providing in-context examples whose query is semantically similar to the target query. Then, we apply standard continued fine-tuning with contrastive loss. We conduct a comprehensive evaluation of new query format across various experimental settings, initializing from both decoder-only checkpoints and pre-trained retriever model checkpoints. We demonstrate that \texttt{RARe} outperforms baseline models across multiple tasks, achieving improvements of up to +1.41\% nDCG@10 compared to methods that do not use in-context examples on standard retrieval benchmarks \citep{thakur2021beir}, and showing even larger gains (+2.72\% nDCG@10) on reasoning-oriented retrieval tasks \citep{xiao2024rarbreasoningretrievalbenchmark}.\looseness=-1

\begin{figure*}
  \centering
  \includegraphics[width=\linewidth]{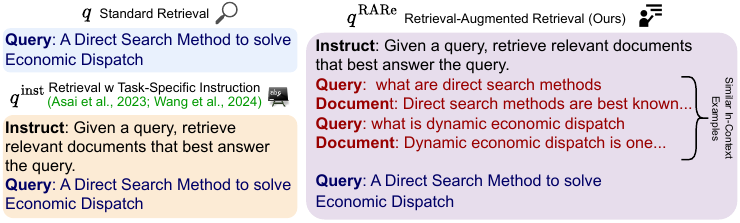}
  \caption{Overview. Prior work augments a task-specific instruction to a given query as input to the Retriever. In \texttt{RARe}, we further leverage a set of in-context exemplars that contain pairs of queries and relevant documents. These in-context examples are augmented with the original query as input to the retriever along with the instruction.
  }
  \label{fig:main}
\end{figure*}

Our contributions can be summarized as follows:
\begin{itemize}[noitemsep,leftmargin=10px]
    \item We introduce \texttt{RARe}, an approach that adapts pre-trained checkpoints to use semantically similar in-context examples for retrieval. 
    \item We demonstrate that this recipe can improve the performance of various base architectures, including decoder-only models and existing retriever models. 
    \item We provide detailed analyses on how the quality, quantity, and selection of in-context examples affect performance, contextualizing the sources of performance gains in embedding-based ICL.
\end{itemize}
All our code and model checkpoints are publicly released.\footnote{\url{https://github.com/atutej/RARe}} 

\section{Method}

\paragraph{Standard Retrieval Setting} We consider a dense retriever ~\citep{karpukhin-etal-2020-dense}, where input queries $q$ and documents $d$ are encoded with an embedder $E(\cdot)$ into a fixed-dimensional embedding. The embedder $E(\cdot)$ is trained on a training set $\train$ which consists of multiple retrieval tasks $\{\train_1, \train_2, \cdots, \train_T\}$, where each task contains training examples of the form $(q, d^+, d^-)$ \citep{wang2024improving, behnamghader2024llm2vec}. Here, $q$ is the input query, $d^+$ is a positive (relevant) document, and $d^-$ is a hard-negative (irrelevant) document, which allows for a contrastive-loss based training.\looseness=-1

The evaluation task $\test$ consists of a corpus of documents $C$, as well as test pairs $(q, R^+)$, where $R^+ = \{d^+_1, d^+_2, ..., d^+_m\} \subset C$ is a set of relevant document(s) for the query \citep{thakur2021beir}. The aim is to retrieve these relevant documents $R^+$ from the corpus $C$ using the embedder $E(\cdot)$. Specifically, an index $C_\ve$ of the corpus with document embeddings $E(d), \forall d \in C$ is created. Then, the embedding $E(q)$ of a test query $q$ is used to retrieve the documents $d$ whose embedding $E(d)$ is closest to $E(q)$, typically with the cosine (cos) similarity function.

\paragraph{Retrieval with Instruction Setting} Current architectures \citep{asai-etal-2023-task, behnamghader2024llm2vec} prepend task-specific instruction $t_i$, $i \in [1, 2, \cdots, T]$ to the query to contextualize the task:\looseness=-1

\begin{equation}
q^{\text{inst}} = \text{Instruct: } \{t_i\} ; \text{ Query: } \{q\}, \quad q \in \train_i 
\label{eq:query_base}
\end{equation}

Then, the embedder $E(\cdot)$ is trained with a standard contrastive loss \citep{izacard2022unsupervised, karpukhin-etal-2020-dense}, incorporating $q^{\text{inst}}$, and $d^+, d^- \in \train_i$, along in-batch negatives $n \in \mathbf{N}$, where $\mathbf{N}$ represents the set of in-batch negatives,

\begin{align}
\ve_{q^{\text{inst}}} = E(q^{\text{inst}}); \quad \ve_{d^+} = E(d^+); \quad \ve_{d^-} = E(d^-); \quad \ve_{n} = E(n) \\
\mathcal{L} = -\log\frac{f(e_{q^{\text{inst}}}, e_{d^+}) }{f(e_{q^{\text{inst}}}, e_{d^+}) + f(e_{q^{\text{inst}}}, e_{d^-}) + \sum_{n \in \mathbf{N}} f(e_{q^{\text{inst}}}, e_n)} \label{eq:loss}
\end{align}

Where $f(x,y)=\exp(\text{cos}(x, y))$. During evaluation on $\test$, each test query is augmented with task-specific instruction $t_\text{test}$.\looseness=-1

\paragraph{BGE-ICL Setting} Recently, \cite{li2024makingtextembeddersfewshot} proposed BGE-ICL, a model that augments in-context examples to enhance the query representation of retrievers. Specifically, given a query $q$, an in-context example set $\train^{\text{ic}}_i = \{(q^{\text{ic}}_1, d^{\text{ic}+}_1), (q^{\text{ic}}_2, d^{\text{ic}+}_2), \cdots, (q^{\text{ic}}_k, d^{\text{ic}+}_k)\}$ is constructed by \textit{randomly sampling} from $\train_i \in \train$, which are then augmented into the original query $q^{\text{inst}}$ to obtain $q^{\text{inst+ic}}$

{
\begin{equation}
\begin{aligned}
\!\!q^{\text{inst+ic}} = \, ``&\text{Instruct: } \{t_i\} ; \text{ Query: } \{q^\text{ic}_1\} ; \text{ Document: } \{d^{\text{ic}+}_1\} \, \cdots; \text{ Query: } \{q\}"
\label{ref:query_format_ic}
\end{aligned}
\end{equation}
} 

Then, $E(\cdot)$ is trained with the same loss as \autoref{eq:loss}, but with $q^{\text{inst+ic}}$ instead of $q^{\text{inst}}$,
\begin{equation}
\mathcal{L}_{\text{in-context}} = -\log\frac{f(e_{q^{\text{inst+ic}}}, e_{d^+}) }{f(e_{q^{\text{inst+ic}}}, e_{d^+}) + f(e_{q^{\text{inst+ic}}}, e_{d^-}) + \sum_{n \in \mathbf{N}} f(e_{q^{\text{inst+ic}}}, e_n)} \label{eq:loss1}
\end{equation}

\begin{algorithm*}[t]
\caption{\texttt{RARe} - Training}
\label{algorithm:RARe}
\KwIn{Training set $\train$, embedder $E(\cdot)$, BM25, the number of in-context examples $k$, mini-batch size $B$.}

\begin{algorithmic}[1]
\FOR{each training iteration}
    \STATE{Sample mini-batch $\mathcal{B}$ of size $B$ from $\train$}
    
    \FOR{$(t_i, q, d^+, d^-) \in \mathcal{B}$}
        \STATE{In-Context Example Retrieval:}
        \STATE $\{q^{\text{RARe}}_1, q^{\text{RARe}}_2, \ldots, q^{\text{RARe}}_k\} \leftarrow \text{Retrieve nearest queries of } q   \text{ from } \train \text{ using BM25}$ 
        \STATE $\{d^{\text{RARe}+}_1, d^{\text{RARe}+}_2, \ldots, d^{\text{RARe}+}_k\} \leftarrow \{d^+ : (q', d^+) \in \train, q' \in \{q^{\text{RARe}}_1, \ldots, q^{\text{RARe}}_k\}\}$
        \STATE $\train^{\text{RARe}}_i \leftarrow \{(q^{\text{RARe}}_1, d^{\text{RARe}+}_1), \ldots, (q^{\text{RARe}}_k, d^{\text{RARe}+}_k)\}$
        \STATE{Query Augmentation:} 
        \STATE $q^{\text{RARe}} = \text{Instruct: } \{t_i\}; \text{ Query: } \{q^{\text{RARe}}_1\}; \text{ Document: } \{d^{\text{RARe}+}_1\} \, \cdots ; \text{ Query: } \{q\}$
    \ENDFOR
    \STATE{Training with Contrastive Loss:}
        \STATE Compute the mini-batch contrastive loss $\mathcal{L}_{\text{in-context}}$ as described in \autoref{eq:loss1}.
        \STATE Update $E(\cdot)$ by minimizing $\mathcal{L}_{\text{in-context}}$.

\ENDFOR
\end{algorithmic}
\KwOut{Trained embedder $E(\cdot)$}
\end{algorithm*}

\paragraph{Our RARe Setting} We propose to enhance the query representation by incorporating \textit{semantically-similar} in-context examples. This provides additional, highly relevant query-specific guidance to the model.

Given a query $q$, we use BM25 \citep{bm25}, a sparse retriever that ranks documents based on keyword matching, and find $k$ closest queries $q_j$ from $\train_i \in \train$ to obtain in-context examples $\train^{\text{RARe}}_i = \{(q^{\text{RARe}}_1, d^{\text{RARe}+}_1), (q^{\text{RARe}}_2, d^{\text{RARe}+}_2), \cdots, (q^{\text{RARe}}_k, d^{\text{RARe}+}_k)\}$. As shown in \autoref{fig:main}, we augment these examples to the original query $q^{\text{inst}}$ to obtain $q^{\text{RARe}}$,
{
\begin{equation}
\begin{aligned}
\!\!q^{\text{RARe}} = \, ``&\text{Instruct: } \{t_i\} ; \text{ Query: } \{q^\text{RARe}_1\} ; \text{ Document: } \{d^{\text{RARe}+}_1\} \, \cdots; \text{ Query: } \{q\}"
\label{ref:query_format_rare}
\end{aligned}
\end{equation}
}

\noindent We then train $E(\cdot)$ using \Cref{eq:loss1} to leverage these semantically similar examples. \Cref{algorithm:RARe} presents our training procedure in detail. At inference time, we similarly perform a search to find nearest in-context examples to form an augmented query. Algorithm \ref{algorithm:RARetest} in the Appendix provides an overview of the inference procedure.

\section{Experimental Setup}
\subsection{Fine-Tuning}
\paragraph{Base Models} We explore two training setups: fine-tuning decoder-only models for retrieval, and fine-tuning existing retriever models. For the first setup, we train the \textit{Llama-3} family of models, following the training methodology outlined by \citet{ma2023finetuningllamamultistagetext, weller2024promptriever}. For the second setup, we use two high-performing publicly available embedding models that were trained with task-specific instructions: \textit{LLM2Vec-Llama-3-8b-Supervised} \citep{behnamghader2024llm2vec} and \textit{E5-Mistral-Instruct} \citep{wang2024improving}. We chose these two models because, unlike some other strong performers \citep{SFRAIResearch2024, moreira2024nvretrieverimprovingtextembedding}, they were not trained on most of the datasets used in our downstream benchmarks. The \textit{LLM2Vec-Llama-3-8b-Supervised} model is initially trained using an unsupervised text reconstruction objective and then fine-tuned with supervised contrastive learning on a public subset of the E5 dataset, which incorporates various supervised training datasets \citep{gao-etal-2021-simcse, msmarco, kwiatkowski-etal-2019-natural}. In contrast, \textit{E5-Mistral-Instruct} undergoes further training on synthetic data that is not publicly available. These models are chosen to assess the impact of additional supervised training on an existing retriever model versus training a model for retrieval from scratch.

\paragraph{Training Data} For fine-tuning existing retriever models, we follow prior work \citep{behnamghader2024llm2vec} and train on a publicly available portion of E5 dataset \citep{springer2024repetition, wang2024improving}, which contains MS-MARCO \citep{msmarco} NLI \citep{gao-etal-2021-simcse}, ELI5 \citep{fan-etal-2019-eli5}, FEVER \citep{thorne-etal-2018-fever}, HotpotQA \citep{yang-etal-2018-hotpotqa}, NQ \citep{kwiatkowski-etal-2019-natural}, SQuAD \citep{rajpurkar-etal-2016-squad}, Quora Duplication Questions \citep{quora-question-pairs}, TriviaQA \citep{joshi-etal-2017-triviaqa}. For fine-tuning models from LLM checkpoint, we use the MS-MARCO \citep{msmarco} passage ranking dataset and train without an instruction prefix, following \citet{ma2023finetuningllamamultistagetext}.

\paragraph{Constructing In-Context Examples} During training, we provide each training example with five in-context examples from the dataset that it belongs to ($k$=5). Specifically, the set of examples $\train^{\text{RARe}}_i$ for each task is drawn from the training set $\train_i$, $q \notin \train^{\text{RARe}}_i$.

\subsection{Evaluation}

\paragraph{Datasets} We evaluate on the widely used BeIR retrieval benchmark \citep{thakur2021beir}. For ablative experiments, we follow prior work and focus on low-resource datasets \citep{wang-etal-2023-query2doc} that potentially benefit more from few-shot examples. Since the BeIR benchmark contains a few datasets whose training sets are in the E5 dataset mixutre, we categorize them as in-domain and out-of-domain (i.e. datasets not seen during training). See \autoref{table:beir_appdx} in the Appendix for a list of in-domain and out-of-domain datasets from BeIR.
We also evaluate on a subset of the RAR-b \citep{xiao2024rarbreasoningretrievalbenchmark} benchmark, which requires complex reasoning for retrievers. Specifically, we evaluate on HellaSwag~\citep{zellers-etal-2019-hellaswag}, PIQA~\citep{bisk2020piqa}, ARC-C~\citep{clark2018think}, TempReason-L1~\citep{tan2023towards}, WinoGrande~\citep{sakaguchi2021winogrande}, $\alpha$-NLI~\citep{Bhagavatula2020Abductive}, SiQA~\citep{sap2019social}, and Quail~\citep{rogers2020getting}. Unlike BeIR, some RAR-b queries are 
composed of sentences with (multiple) indicators (e.g., Start:, End:). Each dataset is associated with a task-specific instruction, following prior work \citep{muennighoff-etal-2023-mteb, wang2024improving, behnamghader2024llm2vec}. We provide additional preprocessing details in \autoref{appdx:experiment_details}.

\paragraph{Constructing In-Context Examples} We construct $\train^{\text{RARe}}_\text{test}$ from the training/development set of each datasets. For datasets on BeIR that do not have either of these, we use a synthetically generated collection of document-query pairs (GenQ) from \citet{thakur2021beir}. For all experiments, we use $k$=5 in-context examples.

\paragraph{Metrics} We use standard metrics for retrieval benchmarks. Following~\citet{thakur2021beir}, we report nDCG@10, which measures the ranking quality of the top 10 retrieved documents, taking into account both the relevance and position of each retrieved document.\looseness=-1

\section{Results}
\label{section:results}

We evaluate in-context example augmented queries in three settings. First, we evaluate the performance after inference-only modification, where we take existing pre-trained retrievers and simply provide in-context examples at inference time (Section \ref{section:results}). Second, we evaluate training retriever with in-context examples from an LLM (decoder-only) backbone (Section \ref{section:llm_checkpoints}). Third, we compare training retriever models with in-context examples from a pre-trained retriever (Section \ref{section:retriever_checkpoints}).\looseness=-1

\begin{figure*}[t]
  \centering
  \includegraphics[width=\linewidth]{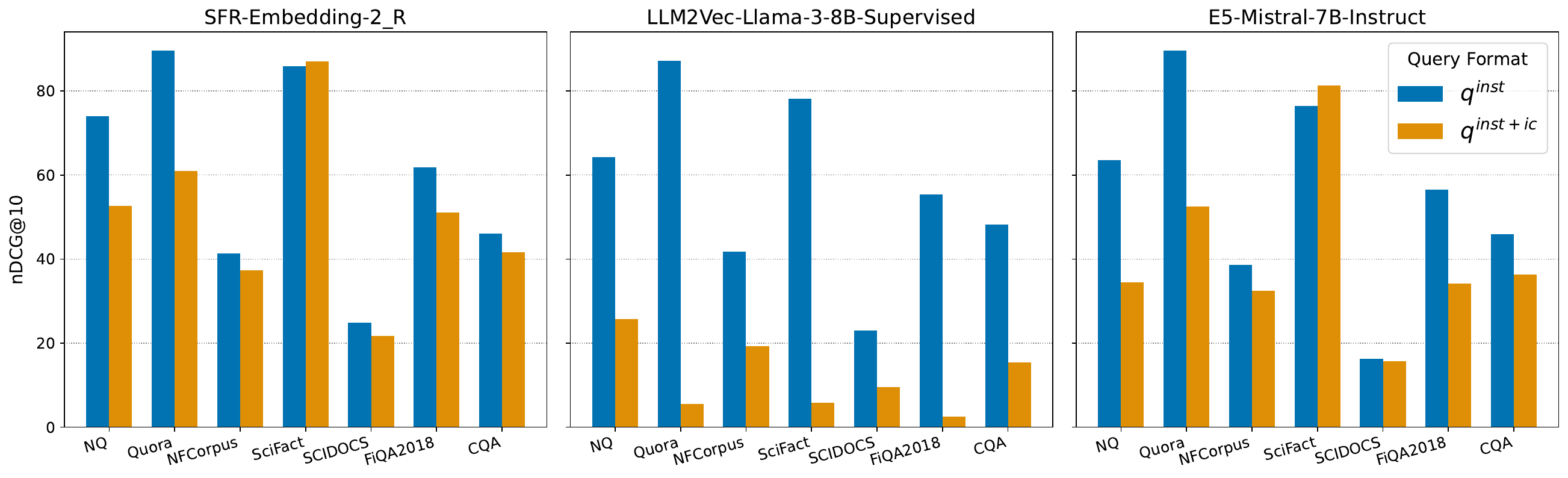}\vspace{-0.5em}
  \caption{\textbf{Inference-only modification does not work.} We report performance before and after adding in-context examples to the query without updating model parameters. Embedding models are not able to leverage in-context examples out of the box, as opposed to decoder-only models.}
  \label{fig:ic_existing_models}
\end{figure*}

\paragraph{Inference-only Modification}
\autoref{fig:ic_existing_models} illustrates the impact of incorporating in-context examples at inference time. Here, we simply modify the query format with retrieved in-context examples (i.e. $q^{\text{RARe}}$, Eq.~\ref{ref:query_format_rare}) at inference time and compare its performance with the query format that does not have retrieved in-context examples (i.e. $q^{\text{inst}}$, Eq.~\ref{eq:query_base}). We evaluate the performance on three retriever models: \textit{SFR-Embedding-2-R} \citep{SFRAIResearch2024}, \textit{LLM2Vec-Llama-3-8B-Supervised} \citep{behnamghader2024llm2vec}, and \textit{E5-Mistral-7B-Instruct} \citep{wang2024improving}. Unlike in autoregressive LLMs, these embedding models generally exhibit decreased performance when in-context examples are added, with \textit{LLM2Vec-Llama-3-8B-Supervised} showing the largest drops in performance, except on one dataset (SciFact), where 2 out of 3 models show marginal gains over providing only instructions. Our experiments, which include adding more in-context examples and using nearest-neighbor examples, extend the findings of \citet{muennighoff2024gritlm}, where in-context examples led to decrease in performance on the GritLM models.

\begin{table*}[t]
\small
\centering
\addtolength{\tabcolsep}{-0pt}
\renewcommand{\arraystretch}{1.1}
\caption{\textbf{Training from decoder-only (LLM) checkpoint.} We initialize all models with \textit{Llama-3.1-8b-Instruct}. Performance is measured by nDCG@10. RARe shows +2.72\% and +0.87\% nDCG@10 over Promptriever and ICL, respectively, on the reasoning oriented RAR-b benchmark. We provide a breakdown of In-Domain (ID), Out-of-Domain (OOD), and overall (Average) performance.}
\begin{tabular}{lcccccc}
\toprule
\multirow{2}{*}{\textbf{Method}} & \multirow{2}{*}{\textbf{Training Data}} &\multicolumn{1}{c}{\textbf{ID}} &\multicolumn{2}{c}{\textbf{OOD}} & \multirow{2}{*}{\textbf{Average}} \\ \cmidrule(lr){3-3} \cmidrule(lr){4-5}
&& BeIR & BeIR & RAR-b & \\ \midrule
RepLLaMA & MS-MARCO & \textbf{43.67} & 54.34 & 19.20 & 39.07 \\
Promptriever  &  MS-MARCO + Synthetic & 42.70& \textbf{56.10}& 20.95 & \underline{39.94}\\
ICL &  MS-MARCO & 42.57 & 53.91 & \underline{22.80} & 39.76 \\
\rowcolor[HTML]{EFEFEF} {\texttt{RARe}} & MS-MARCO & \underline{42.93} & \underline{56.05} & \textbf{23.67} & \textbf{40.88}\\ 
\bottomrule
\end{tabular}
\label{table:repllama_comp}
\end{table*}
\subsection{Training from LLM Checkpoints} 
\label{section:llm_checkpoints}


Next, we present the results of applying our approach when training from LLM checkpoint. This might preserve in-context learning capacity of the LLM, which can be lost during standard IR training, which compresses query and passage into a fixed dimensional vector. We initialize our models with \textit{Llama-3.1-8b-Instruct} \citep{dubey2024llama3herdmodels} to enable comparison with prior work~\cite{ma2023finetuningllamamultistagetext,weller2024promptriever}.\looseness=-1

\paragraph{Comparison Systems} We compare training with our in-context example augmented query with two baselines. The first baseline is vanilla query (Eq.~\ref{eq:query_base}), which was explored in {RepLLaMA} \citep{ma2023finetuningllamamultistagetext}. The second baseline is {Promptriever} \citep{weller2024promptriever} which augments query-specific instructions using a synthetically generated training set from MS-MARCO. The third baseline is ICL, which prepends random in-context examples to the query during training and inference, equivalent to that of \cite{li2024makingtextembeddersfewshot}. In all these systems, the task-specific instruction is a null string \citep{ma2023finetuningllamamultistagetext} as we train on a single task (MS-MARCO).
\paragraph{Results} \autoref{table:repllama_comp} presents the performance on downstream benchmarks when training from LLM checkpoints. Comparing within the same base LLM checkpoint, our apporach outperforms both baselines (RepLLaMA and Promptriever). Our performance is competitive to that of Promptriever \citep{weller2024promptriever}, without incorporating synthetic data during training. Specifically, \texttt{RARe} achieves +0.94\% nDCG@10 over Promptriever, and +1.12\% nDCG@10 over random ICL on average.

\begin{table*}
\small
\centering
\addtolength{\tabcolsep}{-0pt}
\renewcommand{\arraystretch}{1.2}
\caption{\textbf{Training from retriever checkpoints.} {Performance (nDCG@10) on BeIR~\citep{thakur2021beir} and RAR-b~\citep{xiao2024rarbreasoningretrievalbenchmark} benchmarks when fine-tuning retriever model on E5 dataset.} We report a breakdown of performance on In-Domain (ID), Out-of-Domain (OOD, and overall (Average) performance.}
\begin{tabular}{lcccccccc}
\toprule
\multirow{3}{*}{\textbf{Method}} & \multicolumn{4}{c}{\textit{LLM2Vec-Llama-3-8b-Supervised}}  & \multicolumn{4}{c}{\textit{E5-Mistral-Instruct}} \\  \cmidrule(lr){2-5} \cmidrule(lr){6-9}
& \multicolumn{1}{c}{\textbf{ID}} & \multicolumn{2}{c}{\textbf{OOD}} & \multirow{2}{*}{\textbf{Average}} & \multicolumn{1}{c}{\textbf{ID}} & \multicolumn{2}{c}{\textbf{OOD}} & \multirow{2}{*}{\textbf{Average}} \\ \cmidrule(lr){2-2} \cmidrule(lr){3-4} \cmidrule(lr){6-6} \cmidrule(lr){7-8}
 & BeIR & BeIR & RAR-b & & BeIR & BeIR & RAR-b \\ \midrule
Base  & 71.31 &	\underline{49.28} &	21.55 & \underline{47.38} & 71.95 & 49.33 & 22.17 & 47.81 \\
Instruct  & 70.46 & 47.79 & \textbf{23.44} & 47.23 & \underline{72.91} & 48.98 & 24.12 & 48.67 \\
ICL & \textbf{71.81} & 47.69 & 21.94 & 47.14 & 72.03 & \underline{49.46} & \underline{24.69} & \underline{48.72} \\
\rowcolor[HTML]{EFEFEF} {\cellcolor[HTML]{EFEFEF}\texttt{RARe}} &  \underline{71.67} & \textbf{49.30} & \underline{23.10} & \textbf{48.02} & \textbf{72.98} & \textbf{50.93} & \textbf{25.79} & \textbf{49.90} \\  
\bottomrule
\end{tabular}
\label{table:mteb}
\end{table*}

\subsection{Training from Retriever Checkpoints}
\label{section:retriever_checkpoints}

Lastly, we continue training retriever models -- \textit{LLM2Vec-Llama-3-8B-Supervised} \citep{behnamghader2024llm2vec}, \textit{E5-Mistral-Instruct} \citep{wang2024improving} on a training set where queries are augmented with in-context examples. As these initial checkpoints have already been trained on the training dataset, the extent that retrievers adapt to new query format can be limited.\looseness=-1 
\paragraph{Comparison Systems} We first report the initial retriever performance (\textbf{Base}) without any modification. Then, we compare continued fine-tuning with the task-specific instruction query format (Eq.~\ref{eq:query_base}) which only prepends the task specific instruction (\textbf{Instruct}, $q^\text{inst}$). Finally, we compare against \textbf{ICL} \citep{li2024makingtextembeddersfewshot}, which augments random in-context examples to the query ($q^\text{inst+ic}$, Eq.~\ref{ref:query_format_ic}) during training and evaluation.

\paragraph{Results}

\autoref{table:mteb} reports experimental results in this setting. Overall, both fine-tuning approaches provides gains over the base checkpoints. Comparing two settings, Instruct ($q^\text{inst}$) vs. \texttt{RARe} ($q^\text{RARe}$), our method achieves notable improvement with \textit{E5-Mistral-Instruct} base model (up to +1.47\% over random ICL on out-of-domain tasks, and +1.18\% overall). Both RARe and ICL perform similar to Instruct ($q^\text{inst}$) setting when trained with the LLM2Vec base model. It is hard to attribute why experimental results varies based on the base retriever checkpoint, but we note the following differences between the two models. \textit{LLM2Vec-Llama-3-8b-Supervised} is the only model in our experiments where further fine-tuning with only instructions led to a decrease in in-domain performance. \textit{E5-Mistral-Instruct} employs causal attention with last token pooling, and trains on a proprietary synthetic dataset, \textit{LLM2Vec-Llama-3-8b-Supervised} uses bidirectional attention with mean pooling, training only on the public portion. 

The effectiveness of learning with in-context examples may depend on factors such as the underlying model architecture and data setting. Future work can further investigate the impact of base model characteristics, including architectural choices and other constraints.\looseness=-1

\section{Discussions and Analysis}

\begin{figure*}
  \centering
  \includegraphics[width=\linewidth]{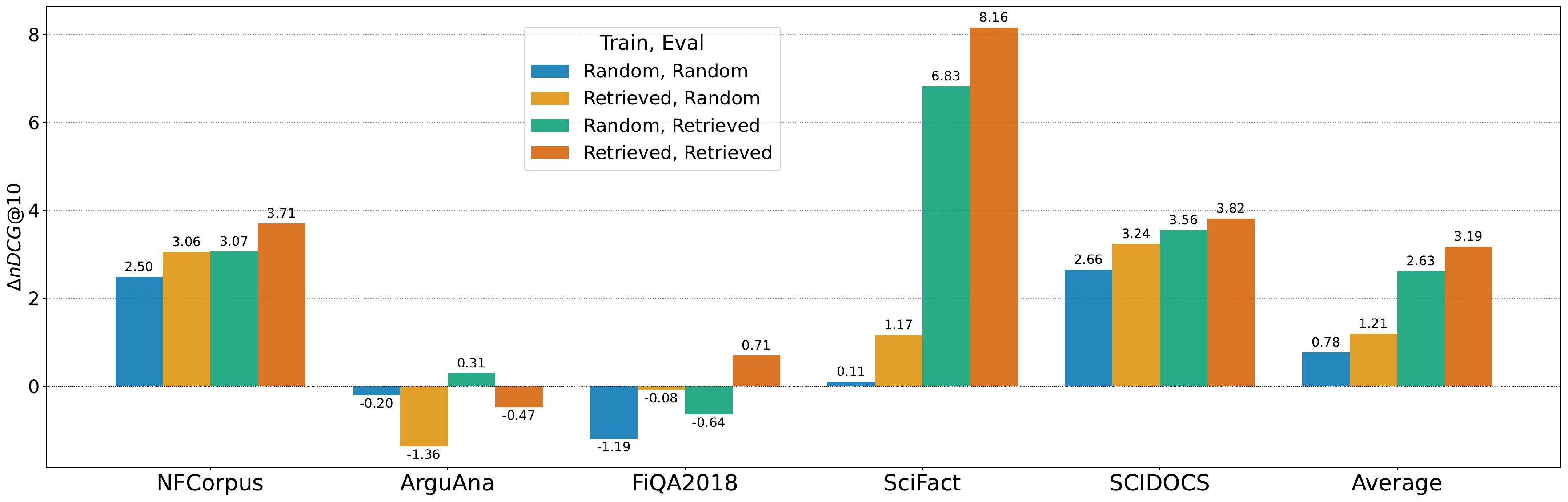}
  \caption{\textbf{Retrieved vs. Random In-context Examples.} Change in performance ($\Delta$nDCG@10) on \textit{E5-Mistral-Instruct} with \texttt{RARe} ($q^\text{RARe}$) from the baseline setting ($q^\text{inst}$ both during training and evaluation time). Using retrieved examples during training and inference enhance model performance in most benchmark datasets.}
  \label{fig:retvsrandom}
\end{figure*}
\subsection{Choice of In-context Examples}
\paragraph{Retrieved (Similar) vs. Random In-Context Examples} In \autoref{fig:retvsrandom}, we study the impact of retrieving the nearest neighbor query-document pairs as examples against randomly chosen examples during training and evaluation. We observe that using retrieved examples during both training and evaluation (Retrieved, Retrieved) consistently outperforms other configurations across most datasets. (Random, Retrieved) and (Retrieved, Random) generally outperform (Random, Random), suggesting retrieved examples are advantageous even when trained with randomly paired in-context examples. On ArguAna, we observe that (Retrieved, Random) performs the worst. There is a mismatch in the lengths of the queries used as in-context examples\footnote{\url{https://huggingface.co/datasets/BeIR/arguana-generated-queries}} (which are significantly shorter) versus the actual test queries in this dataset. This mismatch may introduce variability in performance, which has also been observed in decoder-only LLMs \citep{mishra-etal-2022-reframing}. Overall, our findings align with prior work in in-context learning -- that the incorporation of semantically similar examples is beneficial \citep{agrawal2022incontextexamplesselectionmachine, rubin-etal-2022-learning}. We observe similar overall trends on the other OOD datasets on the BeIR benchmark, reported in \autoref{fig:retvsrandom_appdx} in the Appendix.

\begin{figure*}
  \centering
  \includegraphics[width=\linewidth]{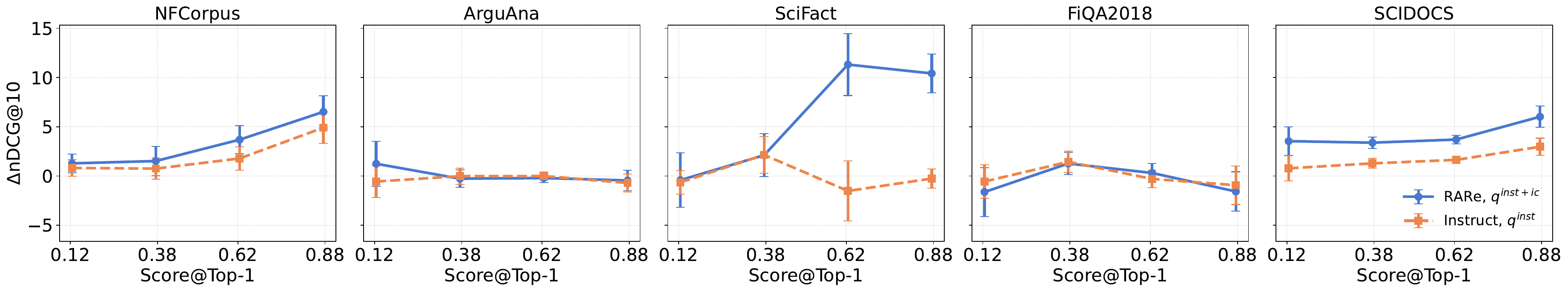}
  \caption{Change in performance ($\Delta$nDCG@10) from the base model (\textit{E5-Mistral-Instruct}) for varying similarity between the closest in-context example query and target query (Score@Top-1). For \texttt{RARe}, we use retrieved in-context examples $q^{\text{RARe}}$ on the augmented in-context query format $q^{\text{inst+ic}}$.}
  \label{fig:top_p_vs_ndcg}
\end{figure*}
\paragraph{Does Having Semantically Relevant In-Context Example Help?} For some test examples, augmented in-context examples are very relevant, and for others, much less so. In this section, we group the evaluation examples by the maximum similarity of in-context query and the test query measured by an off-the-shelf sentence embedding model (Score@Top-1).\footnote{\url{https://huggingface.co/sentence-transformers/all-MiniLM-L6-v2}} and plot the performances for each group. \autoref{fig:top_p_vs_ndcg} presents the performance of our system (\texttt{RARe}) and baseline (Instruct). On NFCorpus and SciFact datasets, we observe that when the closest in-context example has a high similarity with the target query, \texttt{RARe} demonstrates over 10\% gains compared to the base model. On the other hand, fine-tuning with with $q^{\text{inst}}$ exhibits relatively lower performance gains with increasing similarity thresholds. In some datasets, such as ArguAna and FiQA2018, gains with increasing Score@Top-1 are less pronounced, but generally matches the performance of the base model. We observe similar overall trends on other BeIR OOD datasets (\autoref{fig:top_p_vs_ndcg_appdx} in the Appendix).

\begin{table*}
\small
\centering
\addtolength{\tabcolsep}{-2pt}
\renewcommand{\arraystretch}{1.2}
\caption{\textbf{Impact of the number of in-context examples ($k$) during training and evaluation.} All results are on \textit{E5-Mistral-Instruct}. In general, performance increases when increasing the number of examples, and the optimal number of examples depends on the task.}
\begin{tabular}{cccccccc}
\toprule
\textbf{$k$} & \textbf{Arguana} & \textbf{CQADupStack} & \textbf{FiQA2018} & \textbf{NFCorpus} & \textbf{SciFact} & \textbf{Touche2020} & \textbf{Average} \\ \midrule
Instruct (0) & 61.19 & 44.82 & \textbf{57.39} & 40.99 & 77.28 & 29.35 & 51.84 \\ \midrule
1 & 60.47 & 46.76 & 56.07 & 40.67 & 81.47 & \textbf{29.78} & 52.54 \\
3 & \textbf{62.98} & 47.12 & 57.08 & 40.77 & 83.71 & 27.12 & 53.13 \\
\rowcolor[HTML]{EFEFEF} 5 & 60.87 & 48.46 & 57.31 & \textbf{42.28} & 84.79 & 28.70 & 53.74 \\
10 & 58.85 & \textbf{48.92} & 57.03 & 42.24 & \textbf{87.61} & 28.29 & \textbf{53.82} \\ \bottomrule
\end{tabular}
\label{table:n_ic_examples}
\end{table*}
\paragraph{How Many In-Context Examples Are Sufficient?}
We analyze the performance of \text{RARe} when varying the number of in-context examples provided during training and inference. \autoref{table:n_ic_examples} shows that increasing the number of in-context examples generally enhances performance. On ArguAna, we observe that 0 examples are optimal, which is likely due to the mismatch in the lengths of the queries used as in-context examples versus the actual test queries. The impact of adding more in-context example is not uniformly positive across all datasets, suggesting that the optimal number of in-context examples may be dataset-dependent. We observe similar trends when we fix the number of in-context examples to five during training and vary the number of examples provided during inference, which are provided in \autoref{table:eval_n_ic_examples_appdx} in the Appendix.

\begin{table*}
\small
\centering
\addtolength{\tabcolsep}{-3.2pt}
\renewcommand{\arraystretch}{1.2}
\caption{\textbf{In-Context Format} Comparing variants of in-context example format on \textit{E5-Mistral-Instruct}. Instruct refers to the baseline which does not use any in-context examples.}
\begin{tabular}{cccccccc}
\toprule
\textbf{Method} & \textbf{ArguAna} & \textbf{CQA} & \textbf{FiQA2018} & \textbf{NFCorpus} & \textbf{SciFact} & \textbf{Touche2020} & \textbf{Average} \\ \midrule
 Instruct & \textbf{61.19} & 44.82 & \textbf{57.39} & 40.99 & 77.28 & \textbf{29.35} & 51.83 \\ \midrule
 Queries-Only & 58.88 & 46.66 & 54.44 & 41.42 & 78.84 & 28.09 & 51.39 \\
Doc-Only & 57.54 & 48.28 & 56.02 & 41.62 & 79.80 & 29.01 & 52.05 \\
Shuffle-NC & 60.17 & 45.78 & 54.25 & 41.17 & 80.70 & 29.18 & 51.88 \\
 Shuffle-C & 58.97 & 47.97 & 55.98 & 41.78 & 80.51 & 28.97 & 52.36  \\
 \rowcolor[HTML]{EFEFEF} Regular  & 60.87 & \textbf{48.46} & 57.31 & \textbf{42.28} & \textbf{84.79} & 28.70 & \textbf{53.74} \\
\bottomrule
\end{tabular}
\label{table:shuffle}
\end{table*}
\paragraph{Ablating Content and Format of In-context Examples} 
One can view in-context examples as a form of query expansion \citep{lvpseudo, wang-etal-2023-query2doc}, providing useful keywords to improve the performance. In \autoref{table:shuffle}, we analyze the impact of various formats of in-context examples. Each row represents a different model, fine-tuned with the format that they are evaluated on. Query-Only and Doc-Only contain only queries and documents of in-context examples, respectively. For Shuffle-Constrain (C), we randomly shuffle the mapping between $q$ and $d$. On the other hand, for Shuffle-No Constrain (NC), we do not assume any structure, meaning that a query can be followed by a query as well as a document.

First, we observe that Query-Only shows a larger performance drop over Doc-Only, suggesting in-context documents might contain more useful contents than in-context queries. Second, we observe that shuffling the pairings (Shuffle-C) marginally hurts in-context learning in \texttt{RARe}, as opposed to Shuffle-NC. Our findings align with prior study in decoder-only models~\citep{min2022rethinkingroledemonstrationsmakes} which showed strict correspondence between $q$ and $d$ is not required for performance gains from in-context examples. We observe similar trends when keeping the training format fixed (Regular) and vary only the evaluation format (\autoref{table:shuffle_appdx} and \autoref{table:eval_shuffle_appdx} in the Appendix).

\begin{table*}
\centering
\small
\addtolength{\tabcolsep}{-3.9pt}
\renewcommand{\arraystretch}{1.2}
\caption{\textbf{Impact of adding negative documents in the in-context prompt.} All results are on \textit{E5-Mistral-Instruct}. Negative documents $(d^-)$ in the prompt do not enhance performance.}
\begin{tabular}{lccccccc}
\toprule
\textbf{Training / Eval Setting} & \textbf{ArguAna} & \textbf{CQA} & \textbf{FiQA2018} & \textbf{NFCorpus} & \textbf{SciFact} & \textbf{Touche2020} & \textbf{Average} \\ \midrule
{\texttt{RARe}-$q^\text{RARe}$}  & 60.87 & 48.46 & 57.31 & 42.28 & 84.79 & 28.70 & 53.74 \\ \midrule
\texttt{RARe}-$q^\text{RARe+neg}$  & 61.19 & 48.09 & 56.89 & 41.58 & 82.37 & 30.51 & 53.44 \\ \bottomrule
\end{tabular}
\label{table:negatives}
\end{table*}
\paragraph{Negative Documents in the Query} So far, we have used $(q, d^+)$ i.e (Query, Positive Document) pairs as the in-context prompt. Therefore, we study the impact of including negative documents. Specifically, the augmented query $q^{\text{RARe+neg}}$ includes examples of the form $(q, d^+, d^-)$, where the documents are prefixed with the term ``Positive Document: " and ``Negative Document: " respectively. \autoref{table:negatives} presents the downstream performance comparison between \texttt{RARe} variants trained solely on positive examples and those trained with augmented negative documents. The results indicate no performance gains from including negative documents. In fact, training with negative examples led to a slight decrease in performance.

\section{Related Work}

\paragraph{In-context learning} ICL \citep{brown2020language} allows models to adapt to new tasks in a few-shot manner by conditioning on the input data and the context provided at inference time. ICL has been effectively applied to a wide range of tasks such as classification \citep{milios-etal-2023-context}, translation \citep{zhu-etal-2024-multilingual}, mathematical reasoning \citep{wei2022chain, zhou2022teaching}, and code generation \citep{li2023code}. Recent advancements have enhanced the ICL capabilities of language models through additional training procedures \citep{huang2022incontextdistillation, gu-etal-2023-pre, shi2024incontext}. \citet{min-etal-2022-metaicl} and \citet{chen2022metalearning} perform meta-learning with in-context examples on a wide collection of tasks, with the goal of adapting to a new task at inference time through few-shot in-context examples. Other works have explored improving performance through more principled approaches to select in-context examples during inference \citep{zhang-etal-2022-active, sorensen-etal-2022-information, wang2024largelanguage, qin2024incontext, lee2024crafting}. A simple and popular approach is to retrieve examples that are most similar to the input \citep{liu-etal-2022-makes, rubin-etal-2022-learning, li-etal-2023-unified}. Providing in-context examples to re-ranking models has been studied in prior work~\citep{drozdov2023paradepassagerankingusing}, but the potential of augmenting retrievers themselves by leveraging in-context examples remains unexplored. \citet{muennighoff2024gritlm} explored providing an in-context example out-of-the-box, but showed an overall decrease in performance compared to zero-shot inference. The work most closely related to ours is by \citet{li2024makingtextembeddersfewshot}, who use a fixed, randomly selected set of in-context examples rather than retrieving query-specific ones. In contrast, our approach retrieves relevant examples tailored to each query and includes a comprehensive analysis of key design choices such as the quality, quantity, and format of in-context examples, as well as the choice of base model.

\paragraph{Retrieval} Large language models pre-trained with autoregressive  setups \citep{jiang2023mistral7b, dubey2024llama3herdmodels} have shown remarkable performance when adapted to retrieval tasks \citep{wang2024improving, behnamghader2024llm2vec}, outperforming encoder-style retrievers \citep{izacard2022unsupervised, wang2024textembeddings}. Despite these advancements, a challenge that remains is the ability to tailor retrieval systems to specific tasks or queries. To address this, a recent line of work explores incorporating instructions into retrieval by training models to use task-specific instructions along with the query \citep{su-etal-2023-one, asai-etal-2023-task}. \citet{oh2024instructir} and \citet{weller2024followir} further propose using instructions that are specific to each query. Another well-established technique in retrieval is query expansion \citep{jagerman2023queryexpansion, li2023generate, chen-etal-2024-analyze}, where the query is augmented with additional terms to enrich the embedding as a form of relevance feedback \citep{lvpseudo}. Recent efforts have focused on applying LLMs to expand the original query before retrieval \citep{wang-etal-2023-query2doc, shen-etal-2024-retrieval, dai2022promptagatorfewshotdenseretrieval}. However, our focus is on the dense retrieval paradigm where a separate strong generative LLM is not available or used.

\section{Conclusion}
In this paper, we investigated the impact of in-context example augmentation for retrieval models, building on recent efforts to extend in-context learning for retrievers. We introduced \texttt{RARe}, a simple strategy that equips retrievers with the ability to leverage in-context examples by training with semantically similar in-context examples. Through detailed experiments and analyses, we demonstrated that \texttt{RARe} consistently improves performance across various architectures and downstream retrieval tasks, demonstrating the effectiveness of in-context learning for retriever models.

\subsubsection*{Acknowledgments}
The work is partially supported by NSF grant IIS-2312948. This research has been supported by computing support on the Vista GPU Cluster through the Center for Generative AI (CGAI) and the Texas Advanced Computing Center (TACC) at UT Austin. We also thank Fangyuan Xu, Michael Zhang, Anuj Diwan, and other members of the UT NLP community for insightful feedback.

\bibliography{colm2025_conference}

@inproceedings{
thakur2021beir,
title={{BEIR}: A Heterogeneous Benchmark for Zero-shot Evaluation of Information Retrieval Models},
author={Nandan Thakur and Nils Reimers and Andreas R{\"u}ckl{\'e} and Abhishek Srivastava and Iryna Gurevych},
booktitle={Thirty-fifth Conference on Neural Information Processing Systems Datasets and Benchmarks Track (Round 2)},
year={2021},
url={https://openreview.net/forum?id=wCu6T5xFjeJ}
}

@article{
izacard2022unsupervised,
title={Unsupervised Dense Information Retrieval with Contrastive Learning},
author={Gautier Izacard and Mathilde Caron and Lucas Hosseini and Sebastian Riedel and Piotr Bojanowski and Armand Joulin and Edouard Grave},
journal={Transactions on Machine Learning Research},
issn={2835-8856},
year={2022},
url={https://openreview.net/forum?id=jKN1pXi7b0},
note={}
}

@inproceedings{karpukhin-etal-2020-dense,
    title = "Dense Passage Retrieval for Open-Domain Question Answering",
    author = "Karpukhin, Vladimir  and
      Oguz, Barlas  and
      Min, Sewon  and
      Lewis, Patrick  and
      Wu, Ledell  and
      Edunov, Sergey  and
      Chen, Danqi  and
      Yih, Wen-tau",
    booktitle = "Proceedings of the 2020 Conference on Empirical Methods in Natural Language Processing (EMNLP)",
    year = "2020",
    publisher = "Association for Computational Linguistics",
    url = "https://aclanthology.org/2020.emnlp-main.550",
}

@misc{wang2024improving,
      title={Improving Text Embeddings with Large Language Models}, 
      author={Liang Wang and Nan Yang and Xiaolong Huang and Linjun Yang and Rangan Majumder and Furu Wei},
      year={2024},
      eprint={2401.00368},
      archivePrefix={arXiv},
      primaryClass={cs.CL},
      url={https://arxiv.org/abs/2401.00368}, 
}

@misc{springer2024repetition,
      title={Repetition Improves Language Model Embeddings}, 
      author={Jacob Mitchell Springer and Suhas Kotha and Daniel Fried and Graham Neubig and Aditi Raghunathan},
      year={2024},
      eprint={2402.15449},
      archivePrefix={arXiv},
      primaryClass={cs.CL},
      url={https://arxiv.org/abs/2402.15449}, 
}

@misc{behnamghader2024llm2vec,
      title={LLM2Vec: Large Language Models Are Secretly Powerful Text Encoders}, 
      author={Parishad BehnamGhader and Vaibhav Adlakha and Marius Mosbach and Dzmitry Bahdanau and Nicolas Chapados and Siva Reddy},
      year={2024},
      eprint={2404.05961},
      archivePrefix={arXiv},
      primaryClass={cs.CL},
      url={https://arxiv.org/abs/2404.05961}, 
}

@inproceedings{gao-etal-2021-simcse,
    title = {{S}im{CSE}: Simple Contrastive Learning of Sentence Embeddings},
    author = {Gao, Tianyu and Yao, Xingcheng and Chen, Danqi},
    booktitle = {Proceedings of the 2021 Conference on Empirical Methods in Natural Language Processing},
    year = {2021},
    publisher = {Association for Computational Linguistics},
    url = {https://aclanthology.org/2021.emnlp-main.552},
}

@inproceedings{fan-etal-2019-eli5,
    title = "{ELI}5: Long Form Question Answering",
    author = "Fan, Angela  and
      Jernite, Yacine  and
      Perez, Ethan  and
      Grangier, David  and
      Weston, Jason  and
      Auli, Michael",
    booktitle = "Proceedings of the 57th Annual Meeting of the Association for Computational Linguistics",
    month = jul,
    year = "2019",
    publisher = "Association for Computational Linguistics",
    url = "https://aclanthology.org/P19-1346",
}

@inproceedings{thorne-etal-2018-fever,
    title = "{FEVER}: a Large-scale Dataset for Fact Extraction and {VER}ification",
    author = "Thorne, James  and
      Vlachos, Andreas  and
      Christodoulopoulos, Christos  and
      Mittal, Arpit",
    booktitle = "Proceedings of the 2018 Conference of the North {A}merican Chapter of the Association for Computational Linguistics: Human Language Technologies, Volume 1 (Long Papers)",
    month = jun,
    year = "2018",
    publisher = "Association for Computational Linguistics",
    url = "https://aclanthology.org/N18-1074",
}

@article{kwiatkowski-etal-2019-natural,
    title = "Natural Questions: A Benchmark for Question Answering Research",
    author = "Kwiatkowski, Tom  and
      Palomaki, Jennimaria  and
      Redfield, Olivia  and
      Collins, Michael  and
      Parikh, Ankur  and
      Alberti, Chris  and
      Epstein, Danielle  and
      Polosukhin, Illia  and
      Devlin, Jacob  and
      Lee, Kenton  and
      Toutanova, Kristina  and
      Jones, Llion  and
      Kelcey, Matthew  and
      Chang, Ming-Wei  and
      Dai, Andrew M.  and
      Uszkoreit, Jakob  and
      Le, Quoc  and
      Petrov, Slav",
    journal = "Transactions of the Association for Computational Linguistics",
    volume = "7",
    year = "2019",
    address = "Cambridge, MA",
    publisher = "MIT Press",
    url = "https://aclanthology.org/Q19-1026",
}

@inproceedings{rajpurkar-etal-2016-squad,
    title = "{SQ}u{AD}: 100,000+ Questions for Machine Comprehension of Text",
    author = "Rajpurkar, Pranav  and
      Zhang, Jian  and
      Lopyrev, Konstantin  and
      Liang, Percy",
    booktitle = "Proceedings of the 2016 Conference on Empirical Methods in Natural Language Processing",
    month = nov,
    year = "2016",
    publisher = "Association for Computational Linguistics",
    url = "https://aclanthology.org/D16-1264",
}

@inproceedings{joshi-etal-2017-triviaqa,
    title = "{T}rivia{QA}: A Large Scale Distantly Supervised Challenge Dataset for Reading Comprehension",
    author = "Joshi, Mandar  and
      Choi, Eunsol  and
      Weld, Daniel  and
      Zettlemoyer, Luke",
    booktitle = "Proceedings of the 55th Annual Meeting of the Association for Computational Linguistics (Volume 1: Long Papers)",
    month = jul,
    year = "2017",
    publisher = "Association for Computational Linguistics",
    url = "https://aclanthology.org/P17-1147",
}

@misc{quora-question-pairs,
    author = {DataCanary and hilfialkaff and Lili Jiang and Meg Risdal and Nikhil Dandekar and tomtung},
    title = {Quora Question Pairs},
    publisher = {Kaggle},
    year = {2017},
    url = {https://kaggle.com/competitions/quora-question-pairs},
}

@inproceedings{yang-etal-2018-hotpotqa,
    title = "{H}otpot{QA}: A Dataset for Diverse, Explainable Multi-hop Question Answering",
    author = "Yang, Zhilin  and
      Qi, Peng  and
      Zhang, Saizheng  and
      Bengio, Yoshua  and
      Cohen, William  and
      Salakhutdinov, Ruslan  and
      Manning, Christopher D.",
    booktitle = "Proceedings of the 2018 Conference on Empirical Methods in Natural Language Processing",
    year = "2018",
    address = "Brussels, Belgium",
    publisher = "Association for Computational Linguistics",
    url = "https://aclanthology.org/D18-1259",
}

@article{msmarco,
  author    = {Tri Nguyen and
               Mir Rosenberg and
               Xia Song and
               Jianfeng Gao and
               Saurabh Tiwary and
               Rangan Majumder and
               Li Deng},
  title     = {{MS} {MARCO:} {A} Human Generated MAchine Reading COmprehension Dataset},
  journal   = {CoRR},
  volume    = {abs/1611.09268},
  year      = {2016},
  url       = {http://arxiv.org/abs/1611.09268},
  archivePrefix = {arXiv},
  eprint    = {1611.09268},
  bibsource = {dblp computer science bibliography, https://dblp.org}
}

@misc{brown2020language,
      title={Language Models are Few-Shot Learners}, 
      author={Tom B. Brown and Benjamin Mann and Nick Ryder and Melanie Subbiah and Jared Kaplan and Prafulla Dhariwal and Arvind Neelakantan and Pranav Shyam and Girish Sastry and Amanda Askell and Sandhini Agarwal and Ariel Herbert-Voss and Gretchen Krueger and Tom Henighan and Rewon Child and Aditya Ramesh and Daniel M. Ziegler and Jeffrey Wu and Clemens Winter and Christopher Hesse and Mark Chen and Eric Sigler and Mateusz Litwin and Scott Gray and Benjamin Chess and Jack Clark and Christopher Berner and Sam McCandlish and Alec Radford and Ilya Sutskever and Dario Amodei},
      year={2020},
      eprint={2005.14165},
      archivePrefix={arXiv},
      primaryClass={cs.CL},
      url={https://arxiv.org/abs/2005.14165}, 
}

@inproceedings{min-etal-2022-metaicl,
    title = "{M}eta{ICL}: Learning to Learn In Context",
    author = "Min, Sewon  and
      Lewis, Mike  and
      Zettlemoyer, Luke  and
      Hajishirzi, Hannaneh",
    booktitle = "Proceedings of the 2022 Conference of the North American Chapter of the Association for Computational Linguistics: Human Language Technologies",
    month = jul,
    year = "2022",
    address = "Seattle, United States",
    publisher = "Association for Computational Linguistics",
    url = "https://aclanthology.org/2022.naacl-main.201",
}

@misc{chen2022metalearning,
      title={Meta-learning via Language Model In-context Tuning}, 
      author={Yanda Chen and Ruiqi Zhong and Sheng Zha and George Karypis and He He},
      year={2022},
      eprint={2110.07814},
      archivePrefix={arXiv},
      primaryClass={cs.CL},
      url={https://arxiv.org/abs/2110.07814}, 
}

@misc{huang2022incontextdistillation,
      title={In-context Learning Distillation: Transferring Few-shot Learning Ability of Pre-trained Language Models}, 
      author={Yukun Huang and Yanda Chen and Zhou Yu and Kathleen McKeown},
      year={2022},
      eprint={2212.10670},
      archivePrefix={arXiv},
      primaryClass={cs.CL},
      url={https://arxiv.org/abs/2212.10670}, 
}

@misc{shi2024incontext,
      title={In-context Pretraining: Language Modeling Beyond Document Boundaries}, 
      author={Weijia Shi and Sewon Min and Maria Lomeli and Chunting Zhou and Margaret Li and Gergely Szilvasy and Rich James and Xi Victoria Lin and Noah A. Smith and Luke Zettlemoyer and Scott Yih and Mike Lewis},
      year={2024},
      eprint={2310.10638},
      archivePrefix={arXiv},
      primaryClass={cs.CL},
      url={https://arxiv.org/abs/2310.10638}, 
}

@inproceedings{gu-etal-2023-pre,
    title = "Pre-Training to Learn in Context",
    author = "Gu, Yuxian  and
      Dong, Li  and
      Wei, Furu  and
      Huang, Minlie",
    booktitle = "Proceedings of the 61st Annual Meeting of the Association for Computational Linguistics (Volume 1: Long Papers)",
    month = jul,
    year = "2023",
    address = "Toronto, Canada",
    publisher = "Association for Computational Linguistics",
    url = "https://aclanthology.org/2023.acl-long.267",

}

@misc{wang2024largelanguage,
      title={Large Language Models Are Latent Variable Models: Explaining and Finding Good Demonstrations for In-Context Learning}, 
      author={Xinyi Wang and Wanrong Zhu and Michael Saxon and Mark Steyvers and William Yang Wang},
      year={2024},
      eprint={2301.11916},
      archivePrefix={arXiv},
      primaryClass={cs.CL},
      url={https://arxiv.org/abs/2301.11916}, 
}

@inproceedings{liu-etal-2022-makes,
    title = "What Makes Good In-Context Examples for {GPT}-3?",
    author = "Liu, Jiachang  and
      Shen, Dinghan  and
      Zhang, Yizhe  and
      Dolan, Bill  and
      Carin, Lawrence  and
      Chen, Weizhu",
    booktitle = "Proceedings of Deep Learning Inside Out (DeeLIO 2022): The 3rd Workshop on Knowledge Extraction and Integration for Deep Learning Architectures",
    month = may,
    year = "2022",
    address = "Dublin, Ireland and Online",
    publisher = "Association for Computational Linguistics",
    url = "https://aclanthology.org/2022.deelio-1.10",
}

@inproceedings{rubin-etal-2022-learning,
    title = "Learning To Retrieve Prompts for In-Context Learning",
    author = "Rubin, Ohad  and
      Herzig, Jonathan  and
      Berant, Jonathan",
    booktitle = "Proceedings of the 2022 Conference of the North American Chapter of the Association for Computational Linguistics: Human Language Technologies",
    month = jul,
    year = "2022",
    address = "Seattle, United States",
    publisher = "Association for Computational Linguistics",
    url = "https://aclanthology.org/2022.naacl-main.191",
}

@inproceedings{li-etal-2023-unified,
    title = "Unified Demonstration Retriever for In-Context Learning",
    author = "Li, Xiaonan  and
      Lv, Kai  and
      Yan, Hang  and
      Lin, Tianyang  and
      Zhu, Wei  and
      Ni, Yuan  and
      Xie, Guotong  and
      Wang, Xiaoling  and
      Qiu, Xipeng",
    booktitle = "Proceedings of the 61st Annual Meeting of the Association for Computational Linguistics (Volume 1: Long Papers)",
    month = jul,
    year = "2023",
    address = "Toronto, Canada",
    publisher = "Association for Computational Linguistics",
    url = "https://aclanthology.org/2023.acl-long.256",
}

@inproceedings{zhang-etal-2022-active,
    title = "Active Example Selection for In-Context Learning",
    author = "Zhang, Yiming  and
      Feng, Shi  and
      Tan, Chenhao",
    booktitle = "Proceedings of the 2022 Conference on Empirical Methods in Natural Language Processing",
    month = dec,
    year = "2022",
    address = "Abu Dhabi, United Arab Emirates",
    publisher = "Association for Computational Linguistics",
    url = "https://aclanthology.org/2022.emnlp-main.622",
}

@misc{qin2024incontext,
      title={In-Context Learning with Iterative Demonstration Selection}, 
      author={Chengwei Qin and Aston Zhang and Chen Chen and Anirudh Dagar and Wenming Ye},
      year={2024},
      eprint={2310.09881},
      archivePrefix={arXiv},
      primaryClass={cs.CL},
      url={https://arxiv.org/abs/2310.09881}, 
}

@misc{lee2024crafting,
      title={Crafting In-context Examples according to LMs' Parametric Knowledge}, 
      author={Yoonsang Lee and Pranav Atreya and Xi Ye and Eunsol Choi},
      year={2024},
      eprint={2311.09579},
      archivePrefix={arXiv},
      primaryClass={cs.CL},
      url={https://arxiv.org/abs/2311.09579}, 
}

@inproceedings{sorensen-etal-2022-information,
    title = "An Information-theoretic Approach to Prompt Engineering Without Ground Truth Labels",
    author = "Sorensen, Taylor  and
      Robinson, Joshua  and
      Rytting, Christopher  and
      Shaw, Alexander  and
      Rogers, Kyle  and
      Delorey, Alexia  and
      Khalil, Mahmoud  and
      Fulda, Nancy  and
      Wingate, David",
    booktitle = "Proceedings of the 60th Annual Meeting of the Association for Computational Linguistics (Volume 1: Long Papers)",
    month = may,
    year = "2022",
    address = "Dublin, Ireland",
    publisher = "Association for Computational Linguistics",
    url = "https://aclanthology.org/2022.acl-long.60",
}

@misc{oh2024instructir,
      title={INSTRUCTIR: A Benchmark for Instruction Following of Information Retrieval Models}, 
      author={Hanseok Oh and Hyunji Lee and Seonghyeon Ye and Haebin Shin and Hansol Jang and Changwook Jun and Minjoon Seo},
      year={2024},
      eprint={2402.14334},
      archivePrefix={arXiv},
      primaryClass={cs.CL},
      url={https://arxiv.org/abs/2402.14334}, 
}

@misc{weller2024followir,
      title={FollowIR: Evaluating and Teaching Information Retrieval Models to Follow Instructions}, 
      author={Orion Weller and Benjamin Chang and Sean MacAvaney and Kyle Lo and Arman Cohan and Benjamin Van Durme and Dawn Lawrie and Luca Soldaini},
      year={2024},
      eprint={2403.15246},
      archivePrefix={arXiv},
      primaryClass={cs.IR},
      url={https://arxiv.org/abs/2403.15246}, 
}

@inproceedings{asai-etal-2023-task,
    title = "Task-aware Retrieval with Instructions",
    author = "Asai, Akari  and
      Schick, Timo  and
      Lewis, Patrick  and
      Chen, Xilun  and
      Izacard, Gautier  and
      Riedel, Sebastian  and
      Hajishirzi, Hannaneh  and
      Yih, Wen-tau",
    booktitle = "Findings of the Association for Computational Linguistics: ACL 2023",
    month = jul,
    year = "2023",
    address = "Toronto, Canada",
    publisher = "Association for Computational Linguistics",
    url = "https://aclanthology.org/2023.findings-acl.225",
}

@inproceedings{su-etal-2023-one,
    title = "One Embedder, Any Task: Instruction-Finetuned Text Embeddings",
    author = "Su, Hongjin  and
      Shi, Weijia  and
      Kasai, Jungo  and
      Wang, Yizhong  and
      Hu, Yushi  and
      Ostendorf, Mari  and
      Yih, Wen-tau  and
      Smith, Noah A.  and
      Zettlemoyer, Luke  and
      Yu, Tao",
    booktitle = "Findings of the Association for Computational Linguistics: ACL 2023",
    month = jul,
    year = "2023",
    address = "Toronto, Canada",
    publisher = "Association for Computational Linguistics",
    url = "https://aclanthology.org/2023.findings-acl.71",
}

@misc{wang2024textembeddings,
      title={Text Embeddings by Weakly-Supervised Contrastive Pre-training}, 
      author={Liang Wang and Nan Yang and Xiaolong Huang and Binxing Jiao and Linjun Yang and Daxin Jiang and Rangan Majumder and Furu Wei},
      year={2024},
      eprint={2212.03533},
      archivePrefix={arXiv},
      primaryClass={cs.CL},
      url={https://arxiv.org/abs/2212.03533}, 
}

@misc{jiang2023mistral7b,
      title={Mistral 7B}, 
      author={Albert Q. Jiang and Alexandre Sablayrolles and Arthur Mensch and Chris Bamford and Devendra Singh Chaplot and Diego de las Casas and Florian Bressand and Gianna Lengyel and Guillaume Lample and Lucile Saulnier and Lélio Renard Lavaud and Marie-Anne Lachaux and Pierre Stock and Teven Le Scao and Thibaut Lavril and Thomas Wang and Timothée Lacroix and William El Sayed},
      year={2023},
      eprint={2310.06825},
      archivePrefix={arXiv},
      primaryClass={cs.CL},
      url={https://arxiv.org/abs/2310.06825}, 
}

@misc{dubey2024llama3herdmodels,
      title={The Llama 3 Herd of Models}, 
      author={Abhimanyu Dubey and Abhinav Jauhri and Abhinav Pandey and Abhishek Kadian and Ahmad Al-Dahle and Aiesha Letman and Akhil Mathur and Alan Schelten and Amy Yang and Angela Fan and Anirudh Goyal and Anthony Hartshorn and Aobo Yang and Archi Mitra and Archie Sravankumar and Artem Korenev and Arthur Hinsvark and Arun Rao and Aston Zhang and Aurelien Rodriguez and Austen Gregerson and Ava Spataru and others},
      year={2024},
      eprint={2407.21783},
      archivePrefix={arXiv},
      primaryClass={cs.AI},
      url={https://arxiv.org/abs/2407.21783}, 
}

@misc{jagerman2023queryexpansion,
      title={Query Expansion by Prompting Large Language Models}, 
      author={Rolf Jagerman and Honglei Zhuang and Zhen Qin and Xuanhui Wang and Michael Bendersky},
      year={2023},
      eprint={2305.03653},
      archivePrefix={arXiv},
      primaryClass={cs.IR},
      url={https://arxiv.org/abs/2305.03653}, 
}

@article{li2023generate,
  title={Generate, filter, and fuse: Query expansion via multi-step keyword generation for zero-shot neural rankers},
  author={Li, Minghan and Zhuang, Honglei and Hui, Kai and Qin, Zhen and Lin, Jimmy and Jagerman, Rolf and Wang, Xuanhui and Bendersky, Michael},
  journal={arXiv preprint arXiv:2311.09175},
  year={2023}
}

@inproceedings{chen-etal-2024-analyze,
    title = "Analyze, Generate and Refine: Query Expansion with {LLM}s for Zero-Shot Open-Domain {QA}",
    author = "Chen, Xinran  and
      Chen, Xuanang  and
      He, Ben  and
      Wen, Tengfei  and
      Sun, Le",
    booktitle = "Findings of the Association for Computational Linguistics ACL 2024",
    month = aug,
    year = "2024",
    address = "Bangkok, Thailand and virtual meeting",
    publisher = "Association for Computational Linguistics",
    url = "https://aclanthology.org/2024.findings-acl.708",
}

@inproceedings{zhu-etal-2024-multilingual,
    title = "Multilingual Machine Translation with Large Language Models: Empirical Results and Analysis",
    author = "Zhu, Wenhao  and
      Liu, Hongyi  and
      Dong, Qingxiu  and
      Xu, Jingjing  and
      Huang, Shujian  and
      Kong, Lingpeng  and
      Chen, Jiajun  and
      Li, Lei",
    booktitle = "Findings of the Association for Computational Linguistics: NAACL 2024",
    month = jun,
    year = "2024",
    address = "Mexico City, Mexico",
    publisher = "Association for Computational Linguistics",
    url = "https://aclanthology.org/2024.findings-naacl.176"
}

@misc{li2023code,
      title={Large Language Model-Aware In-Context Learning for Code Generation}, 
      author={Jia Li and Ge Li and Chongyang Tao and Jia Li and Huangzhao Zhang and Fang Liu and Zhi Jin},
      year={2023},
      eprint={2310.09748},
      archivePrefix={arXiv},
      primaryClass={cs.SE},
      url={https://arxiv.org/abs/2310.09748}, 
}

@misc{zhou2022teaching,
      title={Teaching Algorithmic Reasoning via In-context Learning}, 
      author={Hattie Zhou and Azade Nova and Hugo Larochelle and Aaron Courville and Behnam Neyshabur and Hanie Sedghi},
      year={2022},
      eprint={2211.09066},
      archivePrefix={arXiv},
      primaryClass={cs.LG},
      url={https://arxiv.org/abs/2211.09066}, 
}

@inproceedings{wei2022chain,
 author = {Wei, Jason and Wang, Xuezhi and Schuurmans, Dale and Bosma, Maarten and ichter, brian and Xia, Fei and Chi, Ed and Le, Quoc V and Zhou, Denny},
 booktitle = {Advances in Neural Information Processing Systems},
 title = {Chain-of-Thought Prompting Elicits Reasoning in Large Language Models},
 volume = {35},
 year = {2022}
}

@inproceedings{milios-etal-2023-context,
    title = "In-Context Learning for Text Classification with Many Labels",
    author = "Milios, Aristides  and
      Reddy, Siva  and
      Bahdanau, Dzmitry",
    booktitle = "Proceedings of the 1st GenBench Workshop on (Benchmarking) Generalisation in NLP",
    month = dec,
    year = "2023",
    address = "Singapore",
    publisher = "Association for Computational Linguistics",
    url = "https://aclanthology.org/2023.genbench-1.14"
}

@inproceedings{wang-etal-2023-query2doc,
    title = "Query2doc: Query Expansion with Large Language Models",
    author = "Wang, Liang  and
      Yang, Nan  and
      Wei, Furu",
    booktitle = "Proceedings of the 2023 Conference on Empirical Methods in Natural Language Processing",
    month = dec,
    year = "2023",
    address = "Singapore",
    publisher = "Association for Computational Linguistics",
    url = "https://aclanthology.org/2023.emnlp-main.585",
}

@misc{xiao2024rarbreasoningretrievalbenchmark,
      title={RAR-b: Reasoning as Retrieval Benchmark}, 
      author={Chenghao Xiao and G Thomas Hudson and Noura Al Moubayed},
      year={2024},
      eprint={2404.06347},
      archivePrefix={arXiv},
      primaryClass={cs.CL},
      url={https://arxiv.org/abs/2404.06347}, 
}

@misc{hu2021lora,
      title={LoRA: Low-Rank Adaptation of Large Language Models}, 
      author={Edward J. Hu and Yelong Shen and Phillip Wallis and Zeyuan Allen-Zhu and Yuanzhi Li and Shean Wang and Lu Wang and Weizhu Chen},
      year={2021},
      eprint={2106.09685},
      archivePrefix={arXiv},
      primaryClass={cs.CL},
      url={https://arxiv.org/abs/2106.09685}, 
}

@misc{douze2024faisslibrary,
      title={The Faiss library}, 
      author={Matthijs Douze and Alexandr Guzhva and Chengqi Deng and Jeff Johnson and Gergely Szilvasy and Pierre-Emmanuel Mazaré and Maria Lomeli and Lucas Hosseini and Hervé Jégou},
      year={2024},
      eprint={2401.08281},
      archivePrefix={arXiv},
      primaryClass={cs.LG},
      url={https://arxiv.org/abs/2401.08281}, 
}

@misc{ma2023finetuningllamamultistagetext,
      title={Fine-Tuning LLaMA for Multi-Stage Text Retrieval}, 
      author={Xueguang Ma and Liang Wang and Nan Yang and Furu Wei and Jimmy Lin},
      year={2023},
      eprint={2310.08319},
      archivePrefix={arXiv},
      primaryClass={cs.IR},
      url={https://arxiv.org/abs/2310.08319}, 
}

@article{bm25,
author = {Robertson, Stephen and Zaragoza, Hugo},
title = {The Probabilistic Relevance Framework: BM25 and Beyond},
year = {2009},
issue_date = {April 2009},
publisher = {Now Publishers Inc.},
address = {Hanover, MA, USA},
volume = {3},
number = {4},
issn = {1554-0669},
url = {https://doi.org/10.1561/1500000019},
journal = {Found. Trends Inf. Retr.},
month = {apr},
pages = {333–389},
numpages = {57}
}

@misc{muennighoff2024gritlm,
      title={Generative Representational Instruction Tuning}, 
      author={Niklas Muennighoff and Hongjin Su and Liang Wang and Nan Yang and Furu Wei and Tao Yu and Amanpreet Singh and Douwe Kiela},
      year={2024},
      eprint={2402.09906},
      archivePrefix={arXiv},
      primaryClass={cs.CL},
      url={https://arxiv.org/abs/2402.09906}, 
}

@inproceedings{
Bhagavatula2020Abductive,
title={Abductive Commonsense Reasoning},
author={Chandra Bhagavatula and Ronan Le Bras and Chaitanya Malaviya and Keisuke Sakaguchi and Ari Holtzman and Hannah Rashkin and Doug Downey and Wen-tau Yih and Yejin Choi},
booktitle={International Conference on Learning Representations},
year={2020},
url={https://openreview.net/forum?id=Byg1v1HKDB}
}

@inproceedings{zellers-etal-2019-hellaswag,
    title = "{H}ella{S}wag: Can a Machine Really Finish Your Sentence?",
    author = "Zellers, Rowan  and
      Holtzman, Ari  and
      Bisk, Yonatan  and
      Farhadi, Ali  and
      Choi, Yejin",
    editor = "Korhonen, Anna  and
      Traum, David  and
      M{\`a}rquez, Llu{\'\i}s",
    booktitle = "Proceedings of the 57th Annual Meeting of the Association for Computational Linguistics",
    month = jul,
    year = "2019",
    address = "Florence, Italy",
    publisher = "Association for Computational Linguistics",
    url = "https://aclanthology.org/P19-1472",
    doi = "10.18653/v1/P19-1472",
    pages = "4791--4800",
}

@inproceedings{bisk2020piqa,
  title={Piqa: Reasoning about physical commonsense in natural language},
  author={Bisk, Yonatan and Zellers, Rowan and Gao, Jianfeng and Choi, Yejin and others},
  booktitle={Proceedings of the AAAI conference on artificial intelligence},
  volume={34},
  pages={7432--7439},
  year={2020}
}

@article{clark2018think,
  title={Think you have solved question answering? try arc, the ai2 reasoning challenge},
  author={Clark, Peter and Cowhey, Isaac and Etzioni, Oren and Khot, Tushar and Sabharwal, Ashish and Schoenick, Carissa and Tafjord, Oyvind},
  journal={arXiv preprint arXiv:1803.05457},
  year={2018}
}

@inproceedings{tan2023towards,
  title={Towards Benchmarking and Improving the Temporal Reasoning Capability of Large Language Models},
  author={Tan, Qingyu and Ng, Hwee Tou and Bing, Lidong},
  booktitle={Proceedings of the 61st Annual Meeting of the Association for Computational Linguistics (Volume 1: Long Papers)},
  pages={14820--14835},
  year={2023}
}

@inproceedings{rogers2020getting,
  title={Getting closer to AI complete question answering: A set of prerequisite real tasks},
  author={Rogers, Anna and Kovaleva, Olga and Downey, Matthew and Rumshisky, Anna},
  booktitle={Proceedings of the AAAI conference on artificial intelligence},
  volume={34},
  pages={8722--8731},
  year={2020}
}

@article{sakaguchi2021winogrande,
  title={Winogrande: An adversarial winograd schema challenge at scale},
  author={Sakaguchi, Keisuke and Bras, Ronan Le and Bhagavatula, Chandra and Choi, Yejin},
  journal={Communications of the ACM},
  volume={64},
  number={9},
  pages={99--106},
  year={2021},
  publisher={ACM New York, NY, USA}
}

@inproceedings{sap2019social,
  title={Social IQa: Commonsense Reasoning about Social Interactions},
  author={Sap, Maarten and Rashkin, Hannah and Chen, Derek and Le Bras, Ronan and Choi, Yejin},
  booktitle={Proceedings of the 2019 Conference on Empirical Methods in Natural Language Processing and the 9th International Joint Conference on Natural Language Processing (EMNLP-IJCNLP)},
  pages={4463--4473},
  year={2019}
}

@misc{SFRAIResearch2024,
  title={SFR-Embedding-Mistral:Enhance Text Retrieval with Transfer Learning},
  author={Meng, Rui and Liu, Ye and Rayhan Joty, Shafiq and Xiong, Caiming and Zhou, Yingbo and Yavuz, Semih},
  howpublished={Salesforce AI Research Blog},
  year={2024},
  url={https://blog.salesforceairesearch.com/sfr-embedded-mistral/}
}

@misc{weller2024promptriever,
      title={Promptriever: Instruction-Trained Retrievers Can Be Prompted Like Language Models}, 
      author={Orion Weller and Benjamin Van Durme and Dawn Lawrie and Ashwin Paranjape and Yuhao Zhang and Jack Hessel},
      year={2024},
      eprint={2409.11136},
      archivePrefix={arXiv},
      primaryClass={cs.IR},
      url={https://arxiv.org/abs/2409.11136}, 
}

@misc{lee2024nvembedimprovedtechniquestraining,
      title={NV-Embed: Improved Techniques for Training LLMs as Generalist Embedding Models}, 
      author={Chankyu Lee and Rajarshi Roy and Mengyao Xu and Jonathan Raiman and Mohammad Shoeybi and Bryan Catanzaro and Wei Ping},
      year={2024},
      eprint={2405.17428},
      archivePrefix={arXiv},
      primaryClass={cs.CL},
      url={https://arxiv.org/abs/2405.17428}, 
}

@misc{min2022rethinkingroledemonstrationsmakes,
      title={Rethinking the Role of Demonstrations: What Makes In-Context Learning Work?}, 
      author={Sewon Min and Xinxi Lyu and Ari Holtzman and Mikel Artetxe and Mike Lewis and Hannaneh Hajishirzi and Luke Zettlemoyer},
      year={2022},
      eprint={2202.12837},
      archivePrefix={arXiv},
      primaryClass={cs.CL},
      url={https://arxiv.org/abs/2202.12837}, 
}

@misc{dong2024surveyincontextlearning,
      title={A Survey on In-context Learning}, 
      author={Qingxiu Dong and Lei Li and Damai Dai and Ce Zheng and Jingyuan Ma and Rui Li and Heming Xia and Jingjing Xu and Zhiyong Wu and Tianyu Liu and Baobao Chang and Xu Sun and Lei Li and Zhifang Sui},
      year={2024},
      eprint={2301.00234},
      archivePrefix={arXiv},
      primaryClass={cs.CL},
      url={https://arxiv.org/abs/2301.00234}, 
}

@misc{saadfalcon2024benchmarkingbuildinglongcontextretrieval,
      title={Benchmarking and Building Long-Context Retrieval Models with LoCo and M2-BERT}, 
      author={Jon Saad-Falcon and Daniel Y. Fu and Simran Arora and Neel Guha and Christopher Ré},
      year={2024},
      eprint={2402.07440},
      archivePrefix={arXiv},
      primaryClass={cs.IR},
      url={https://arxiv.org/abs/2402.07440}, 
}

@misc{zhang2024mambaretrieverutilizingmamba,
      title={Mamba Retriever: Utilizing Mamba for Effective and Efficient Dense Retrieval}, 
      author={Hanqi Zhang and Chong Chen and Lang Mei and Qi Liu and Jiaxin Mao},
      year={2024},
      eprint={2408.08066},
      archivePrefix={arXiv},
      primaryClass={cs.IR},
      url={https://arxiv.org/abs/2408.08066}, 
}

@misc{agrawal2022incontextexamplesselectionmachine,
      title={In-context Examples Selection for Machine Translation}, 
      author={Sweta Agrawal and Chunting Zhou and Mike Lewis and Luke Zettlemoyer and Marjan Ghazvininejad},
      year={2022},
      eprint={2212.02437},
      archivePrefix={arXiv},
      primaryClass={cs.CL},
      url={https://arxiv.org/abs/2212.02437}, 
}

@misc{xu2023knnpromptingbeyondcontextlearning,
      title={$k$NN Prompting: Beyond-Context Learning with Calibration-Free Nearest Neighbor Inference}, 
      author={Benfeng Xu and Quan Wang and Zhendong Mao and Yajuan Lyu and Qiaoqiao She and Yongdong Zhang},
      year={2023},
      eprint={2303.13824},
      archivePrefix={arXiv},
      primaryClass={cs.CL},
      url={https://arxiv.org/abs/2303.13824}, 
}

@misc{moreira2024nvretrieverimprovingtextembedding,
      title={NV-Retriever: Improving text embedding models with effective hard-negative mining}, 
      author={Gabriel de Souza P. Moreira and Radek Osmulski and Mengyao Xu and Ronay Ak and Benedikt Schifferer and Even Oldridge},
      year={2024},
      eprint={2407.15831},
      archivePrefix={arXiv},
      primaryClass={cs.IR},
      url={https://arxiv.org/abs/2407.15831}, 
}

@misc{drozdov2023paradepassagerankingusing,
      title={PaRaDe: Passage Ranking using Demonstrations with Large Language Models}, 
      author={Andrew Drozdov and Honglei Zhuang and Zhuyun Dai and Zhen Qin and Razieh Rahimi and Xuanhui Wang and Dana Alon and Mohit Iyyer and Andrew McCallum and Donald Metzler and Kai Hui},
      year={2023},
      eprint={2310.14408},
      archivePrefix={arXiv},
      primaryClass={cs.IR},
      url={https://arxiv.org/abs/2310.14408}, 
}

@inproceedings{muennighoff-etal-2023-mteb,
    title = "{MTEB}: Massive Text Embedding Benchmark",
    author = "Muennighoff, Niklas  and
      Tazi, Nouamane  and
      Magne, Loic  and
      Reimers, Nils",
    editor = "Vlachos, Andreas  and
      Augenstein, Isabelle",
    booktitle = "Proceedings of the 17th Conference of the European Chapter of the Association for Computational Linguistics",
    month = may,
    year = "2023",
    address = "Dubrovnik, Croatia",
    publisher = "Association for Computational Linguistics",
    url = "https://aclanthology.org/2023.eacl-main.148",
    doi = "10.18653/v1/2023.eacl-main.148",
    pages = "2014--2037",
}

@inproceedings{lvpseudo,
author = {Lv, Yuanhua and Zhai, ChengXiang},
title = {A comparative study of methods for estimating query language models with pseudo feedback},
year = {2009},
isbn = {9781605585123},
publisher = {Association for Computing Machinery},
address = {New York, NY, USA},
url = {https://doi.org/10.1145/1645953.1646259},
doi = {10.1145/1645953.1646259},
booktitle = {Proceedings of the 18th ACM Conference on Information and Knowledge Management},
pages = {1895–1898},
numpages = {4},
location = {Hong Kong, China},
series = {CIKM '09}
}

@inproceedings{shen-etal-2024-retrieval,
    title = "Retrieval-Augmented Retrieval: Large Language Models are Strong Zero-Shot Retriever",
    author = "Shen, Tao  and
      Long, Guodong  and
      Geng, Xiubo  and
      Tao, Chongyang  and
      Lei, Yibin  and
      Zhou, Tianyi  and
      Blumenstein, Michael  and
      Jiang, Daxin",
    editor = "Ku, Lun-Wei  and
      Martins, Andre  and
      Srikumar, Vivek",
    booktitle = "Findings of the Association for Computational Linguistics ACL 2024",
    month = aug,
    year = "2024",
    address = "Bangkok, Thailand and virtual meeting",
    publisher = "Association for Computational Linguistics",
    url = "https://aclanthology.org/2024.findings-acl.943",
    doi = "10.18653/v1/2024.findings-acl.943",
    pages = "15933--15946"
}

@misc{li2024makingtextembeddersfewshot,
      title={Making Text Embedders Few-Shot Learners}, 
      author={Chaofan Li and MingHao Qin and Shitao Xiao and Jianlyu Chen and Kun Luo and Yingxia Shao and Defu Lian and Zheng Liu},
      year={2024},
      eprint={2409.15700},
      archivePrefix={arXiv},
      primaryClass={cs.IR},
      url={https://arxiv.org/abs/2409.15700}, 
}

@inproceedings{mishra-etal-2022-reframing,
    title = "Reframing Instructional Prompts to {GPT}k{'}s Language",
    author = "Mishra, Swaroop  and
      Khashabi, Daniel  and
      Baral, Chitta  and
      Choi, Yejin  and
      Hajishirzi, Hannaneh",
    editor = "Muresan, Smaranda  and
      Nakov, Preslav  and
      Villavicencio, Aline",
    booktitle = "Findings of the Association for Computational Linguistics: ACL 2022",
    month = may,
    year = "2022",
    address = "Dublin, Ireland",
    publisher = "Association for Computational Linguistics",
    url = "https://aclanthology.org/2022.findings-acl.50",
    doi = "10.18653/v1/2022.findings-acl.50",
    pages = "589--612"
}

@misc{lei2024corpussteeredqueryexpansionlarge,
      title={Corpus-Steered Query Expansion with Large Language Models}, 
      author={Yibin Lei and Yu Cao and Tianyi Zhou and Tao Shen and Andrew Yates},
      year={2024},
      eprint={2402.18031},
      archivePrefix={arXiv},
      primaryClass={cs.IR},
      url={https://arxiv.org/abs/2402.18031}, 
}

@misc{jedidi2024zeroshotdenseretrievalembeddings,
      title={Zero-Shot Dense Retrieval with Embeddings from Relevance Feedback}, 
      author={Nour Jedidi and Yung-Sung Chuang and Leslie Shing and James Glass},
      year={2024},
      eprint={2410.21242},
      archivePrefix={arXiv},
      primaryClass={cs.IR},
      url={https://arxiv.org/abs/2410.21242}, 
}

@misc{dai2022promptagatorfewshotdenseretrieval,
      title={Promptagator: Few-shot Dense Retrieval From 8 Examples}, 
      author={Zhuyun Dai and Vincent Y. Zhao and Ji Ma and Yi Luan and Jianmo Ni and Jing Lu and Anton Bakalov and Kelvin Guu and Keith B. Hall and Ming-Wei Chang},
      year={2022},
      eprint={2209.11755},
      archivePrefix={arXiv},
      primaryClass={cs.CL},
      url={https://arxiv.org/abs/2209.11755}, 
}
\bibliographystyle{colm2025_conference}

\clearpage
\appendix
\onecolumn
\section*{Appendix}


The appendix is organized as follows: In \autoref{appdx:experiment_details}, we present details on additional data preprocessing and other training details. In \autoref{appdx:additional_experiments}, we present additional results and experiments.\\

\section{Experimental Details}
\label{appdx:experiment_details}

\subsection{Training Details}
\paragraph{Hyperparameters} For fine-tuning \textit{Llama-3-8B}, we follow the setting outlined in \citet{ma2023finetuningllamamultistagetext}. We train on 4 H100 GPUs with per-device batch size 8 and gradient accumulation steps 4. We apply LoRA \citep{hu2021lora} with $r$=32,  temperature of 0.01, learning rate 1e-4 with 100 warmup steps. We use a sequence length of 512 for documents and 1024 for queries as in-context augmented queries are longer. For \texttt{RARe} we use a mixture of 70\% examples with in-context examples and 30\% without \autoref{table:data_mix}.

When fine-tuning existing retriever models (\textit{E5-Mistral-Instruct}, \textit{LLM2Vec-Llama-3-8B-Supervised}), we follow a setting similar to \citet{behnamghader2024llm2vec}. We train on 8 H100 GPUs with a largest possible per-device batch size of 32 along with 2 gradient accumulation steps. We consider a random subset of $100K$ examples from the public E5 dataset mixture \citep{springer2024repetition, wang2024improving}. We use a learning rate of 2e-4, maximum sequence length 1024, warmup ratio 0.1 for 1 epoch. For \textit{E5-Mistral-Instruct}, we apply LoRA \citep{hu2021lora} $r$=16, and $r$=4 for \textit{LLM2Vec-Llama-3-8B-Supervised} since a higher rank was leading to severe overfitting on the instruction baseline.

\subsection{Data Processing}
\paragraph{RAR-b} Since RAR-b benchmark provides only test split, we parse the original training data for each dataset to use as in-context examples. We exclude datasets without any training splits and 2 datasets that were a mixture of multiple tasks or datasets, thereby being difficult to parse. This results in 8 datasets to evaluate on. We preprocess the training split to match the format of RAR-b test split, without excluding any instances. An exception is made for $\alpha$-NLI, where there were multiple identical instances. Therefore, we removed such duplicates, resulting in 72,046 in-context candidates. Furthermore, some RAR-b queries are composed of sentences with (multiple) indicators (e.g., Start:, End:). To address this, we make a minor modification in formatting, enclosing the queries in brackets. The final query representation is:

\begin{equation}
    \begin{aligned}
    q^{\text{RARe}} = \,\, & ``\text{Instruct: } \{t\} ; \text{ Query: } [\{q^\text{RARe}_1\}] ; \text{ Document: } \{d^{\text{RARe}+}_1\} \, \cdots ; &\text{ Query: } [\{q\}]"
    \end{aligned}
\end{equation}

\paragraph{Inference Algorithm} Algorithm \ref{algorithm:RARetest} provides a detailed outline of inference with \texttt{RARe}.

\paragraph{Promptriever}
Promptriever ~\citep{weller2024promptriever} employs 10 different prompts and reports the highest score for each dataset. We apply the prompt that works the best (outperforms 5/15 datasets), which is as follows: \texttt{A document that meets these criteria is considered relevant, while a document that does not meet these criteria is considered non-relevant.}

\paragraph{Synthetic In-Context Examples for BeIR}
For QuoraRetrieval, MSMARCO, DBPedia, FiQA-2018, NFCorpus, and SciFact, we utilize the original training splits without incorporating synthetically generated examples. For the remaining datasets—ArguAna, ClimateFEVER, FEVER, HotpotQA, Touche2020, TREC-COVID, CQADupStack, SCIDOCS, and NQ—we employ synthetic examples provided by the BEIR benchmark (e.g., ArguAna: \url{https://huggingface.co/datasets/BeIR/arguana-generated-queries}). For further details, please refer to our codebase at \url{https://github.com/atutej/RARe}.

\begin{algorithm*}
\caption{\texttt{RARe} - Inference}
\label{algorithm:RARetest}
\KwIn{A list of test queries $D^{test}$, Corpus $C$, embedder $E(\cdot)$, the number of in-context examples $k$, Training dataset $\train^{\mathcal{T}}$, task instruction $t$.}

\begin{algorithmic}[1]
\STATE $C_\ve \leftarrow$ Construct document index as $E(d), \forall d \in C$. 
 \FOR{$i \in [0, len(D^{test})]$}
 \STATE{$q=D^\text{test}[i]$}
 \STATE{In-Context Example Retrieval:}
    \STATE $\{q_{1}^{\text{RARe}}, q_{2}^{\text{RARe}}, \ldots, q_{k}^{\text{RARe}}\}$ $\leftarrow$ Retrieve nearest neighbor queries of $q$ from $\train^{\mathcal{T}}$ using BM25
    \STATE $\{d^{\text{RARe}+}_1, d^{\text{RARe}+}_2, \ldots, d^{\text{RARe}+}_k\} \leftarrow \{d^+ : (q', d^+) \in \train, q' \in \{q^{\text{RARe}}_1, \ldots, q^{\text{RARe}}_k\}\}$
    \STATE $\train^{\text{RARe}} \leftarrow \{(q^{\text{RARe}}_1, d^{\text{RARe}+}_1), \ldots, (q^{\text{RARe}}_k, d^{\text{RARe}+}_k)\}$

 \STATE{Query Augmentation / Encoding:} 
    \STATE $q^{\text{RARe}} = \text{Instruct: } \{t\}; \text{ Query: } \{q^{\text{RARe}}_1\}; \text{ Document: } \{d^{\text{RARe}+}_1\} \, \cdots ; \text{ Query: } \{q\}$ \\
    \STATE{ $\ve_q \leftarrow E(q^{\text{RARe}}_{\text{test}})$}
    
 \STATE{Prediction:}
 
    \STATE{$d = \text{argmax}_{d\in C} \exp(\text{cos}(\ve_{q}, \ve_{d}))$}
    \STATE{$D_\text{pred}$.append($d$)}
\ENDFOR
\end{algorithmic}
\KwOut{Predictions $D_\text{pred}$.}
\end{algorithm*}

\section{Additional Experiments}
\label{appdx:additional_experiments}

\begin{table*}[ht]
\small
\centering
\addtolength{\tabcolsep}{-2pt}
\renewcommand{\arraystretch}{1.1}
\caption{Performance (nDCG@10) on BeIR~\citep{thakur2021beir} when fine-tuning retriever model on E5 dataset. We report a breakdown of performance on In-Domain (ID) and Out-of-Domain (OOD) tasks on BeIR. * indicates statistical significance over Instruct $(p < 0.05)$ using the paired bootstrap test. For the Average score, we compute the overall p-value using Fisher's method.}
\begin{tabular}{llccccccccc}
\toprule
\multicolumn{2}{c}{} & \multicolumn{4}{c}{\textit{LLM2Vec-Llama-3-8b-Supervised}} & \multicolumn{4}{c}{\textit{E5-Mistral-Instruct}} \\ \cmidrule(lr){3-6} \cmidrule(lr){7-10}
\textbf{Category} & \textbf{Dataset} & Base & Instruct & \multicolumn{2}{c}{\texttt{RARe}} & Base & Instruct &\multicolumn{2}{c}{\texttt{RARe}} \\ \cmidrule(lr){3-3} \cmidrule(lr){4-4} \cmidrule(lr){5-6} \cmidrule(lr){7-7} \cmidrule(lr){8-8} \cmidrule(lr){9-10} 
& Eval Format & $q^{\text{inst}}$ & $q^{\text{inst}}$ & $q^{\text{inst}}$ & $q^{\text{RARe}}$ & $q^{\text{inst}}$  & $q^{\text{inst}}$ &  $q^{\text{inst}}$ & $q^{\text{RARe}}$ \\ \midrule
\multirow{5}{*}{ID} 
 & FEVER & \textbf{90.20} & 88.12 & 88.43 & 86.62 & 87.84 & \textbf{91.50} & 90.18 & 90.48 \\
 & HotpotQA & 71.76 & 72.50 & 73.83 & \textbf{79.09}* & 75.72 & 73.91 & 72.18 & \textbf{75.95}* \\
 & NQ & 64.21 & 63.63 & 65.00 & \textbf{66.13}* & 63.53 & 67.44 & \textbf{68.15} & 67.66 \\
 & QuoraRetrieval & 87.16 & 87.85 & \textbf{87.88} & 87.63 & 89.61 & \textbf{89.82} & 89.59 & 88.95 \\
 & MSMARCO & \textbf{43.24} & 40.19 & 40.77 & 38.88 & 43.06 & 41.89 & 41.88 & 41.88 \\ \midrule
\multirow{10}{*}{OOD} 
 & ArguAna & \textbf{62.78} & 60.51 & 59.54 & 57.05 & 61.65 & 61.19 & \textbf{62.90} & 60.87 \\
 & ClimateFEVER & 34.27 & 34.49 & 34.67 & \textbf{34.73}* & 38.35 & \textbf{39.03} & 38.99 & 37.50 \\
 & CQADupStack & 48.25 & 49.76 & 49.10 & \textbf{49.93} & 42.97 & 44.82 & 45.57 & \textbf{48.46}* \\
 & DBPedia & 48.34 & 48.61 & 48.41 & \textbf{49.09}* & 48.89 & 48.92 & 49.24 & \textbf{49.65}* \\
 & FiQA2018 & \textbf{55.33} & 52.99 & 54.26 & 52.82 & 56.81 & \textbf{57.39} & 56.33 & 57.31 \\
 & NFCorpus & 41.83 & \textbf{41.92} & 41.61 & 41.84 & 38.58 & 40.99 & 41.19 & \textbf{42.28}* \\
 & SCIDOCS & 22.96 & \textbf{23.97} & 22.92 & 23.35 & 16.32 & 17.94 & 18.71 & \textbf{20.19}* \\
 & SciFact & 78.22 & 76.89 & 77.70 & \textbf{81.77}* & 76.42 & 77.28 & 77.11 & \textbf{84.79}* \\
 & Touche2020 & 20.50 & 22.11 & \textbf{22.71} & 19.54 & 26.27 & \textbf{29.35} & 27.56 & 28.7 \\
 & TRECCOVID & 80.34 & 68.37 & 78.55 & \textbf{82.78}* & \textbf{87.03} & 72.89 & 77.03 & 79.58 \\ \midrule
 & \textbf{Average} & 56.63 & 55.35 & 56.36 & \textbf{56.76}* & 56.87 & 56.96 & 57.11 & \textbf{58.28}* \\ \bottomrule
\end{tabular}
\label{table:beir_appdx}
\end{table*}

\begin{table*}[ht]
\small
\centering
\caption{Performance on reasoning-focused IR benchmark RAR-b \citep{xiao2024rarbreasoningretrievalbenchmark} when fine-tuning existing retriever models. * indicates statistical significance over Instruct $(p < 0.05)$ using the paired bootstrap test. For the Average score, we compute the overall p-value using Fisher's method. }
\begin{tabular}{lcccccccc}
\toprule
 & \multicolumn{4}{c}{\textit{LLM2Vec-Llama-3-8b-Supervised}} & \multicolumn{4}{c}{\textit{E5-Mistral-Instruct}} \\ \cmidrule(lr){2-5} \cmidrule(lr){6-9}
\textbf{Dataset} & Base & Instruct & \multicolumn{2}{c}{RARe} & Base & Instruct & \multicolumn{2}{c}{RARe} \\ 
Eval Format & $q^{\text{inst}}$ & $q^{\text{inst}}$ & $q^{\text{inst}}$ & $q^{\text{RARe}}$ & $q^{\text{inst}}$ & $q^{\text{inst}}$ & $q^{\text{inst}}$ & $q^{\text{RARe}}$ \\ \midrule 
 ARC-C & \textbf{18.81} & 18.77 & 18.28 & 17.02 & 19.00 & 20.37 & 22.72 & \textbf{26.44}* \\
  $\alpha$-NLI & 26.59 & \textbf{27.29} & 25.25 &  23.66 & \textbf{26.04} & 25.70 & 24.19 & 23.23 \\
  HellaSwag & \textbf{34.32} & 34.19 & 34.19 & 33.29 & 35.38 & 35.99 & 35.07 & \textbf{36.29}* \\
  PIQA & 33.57 & 37.07 & 38.12 & \textbf{39.72}* & 39.80 & 39.35 & 37.22 & \textbf{41.35}* \\
  Quail & \textbf{6.83} & 6.06 & 5.57 & 4.25 & 8.40 & 10.94 & \textbf{15.34} & 14.69 \\
  SiQA & \textbf{6.99} & 5.34 & 4.39 & 4.55 & \textbf{5.66} & 5.45 & 5.75 & \textbf{6.15} \\
  TempReason-L1 & 5.24 & 5.89 & 5.55 & \textbf{7.87*} & 3.60 & \textbf{4.71} & 4.55 & 4.67 \\
  WinoGrande & 40.02 & 52.88 & 48.47 & \textbf{54.44}* & 39.48 & 50.41 & 44.26  & \textbf{53.50}* \\ \midrule
  \textbf{Average} & 21.55 & \textbf{23.44} & 22.48 & 23.10 & 22.17 & 24.12 & 23.64 & \textbf{25.79}* \\ \bottomrule
\end{tabular}
\label{table:rarb_appdx}
\end{table*}

\begin{table*}[ht]
\small
\centering
\caption{Performance (nDCG@10) on BeIR when training decoder-only models. * indicates statistical significance over RepLLaMA $(p < 0.05)$ using the paired bootstrap test. For the Average score, we compute the overall p-value using Fisher's method.}
\addtolength{\tabcolsep}{-3pt}
\begin{tabular}{lccccccc}
\toprule
\multirow{3}{*}{\textbf{Dataset}} & \textit{Llama2} & \multicolumn{2}{c}{\textit{Llama3}} & \multicolumn{4}{c}{\textit{Llama-3.1-Instruct}} \\ \cmidrule(l){2-2} \cmidrule(l){3-4} \cmidrule(l){5-8} 
 & RepLLaMA & RepLLaMA & RARe & \multicolumn{1}{c}{RepLLaMA} & Promptreiver & \multicolumn{2}{c}{RARe} \\
 Eval Format & $q^{\text{inst}}$ & $q^{\text{inst}}$ & $q^{\text{RARe}}$ & \multicolumn{1}{c}{$q^{\text{inst}}$} & $q^{\text{inst}}$ & $q^{\text{inst}}$ & $q^{\text{RARe}}$ \\ \midrule
ArguAna & 48.60 & 52.83 & 49.48 & 51.38 & 58.90 & \textbf{54.77} & 52.83 \\
ClimateFEVER & 29.30 & 32.52 & 32.12 & 33.13 & 29.80 & \textbf{35.91} & 34.24* \\
CQADupStack & 37.91 & 42.59 & 42.96 & 41.58 & 42.18 & 42.55 & \textbf{43.31}* \\
DBPedia & 44.80 & 45.62 & 45.79 & 44.73 & \textbf{46.00} & 45.87 & 45.95* \\
FEVER & 82.90 & 81.79 & 83.66 & 79.22 & \textbf{85.50} & 80.05 & 81.84* \\
FiQA2018 & 45.00 & 44.31 & 47.13 & 44.50 & \textbf{47.20} & 44.36 & 46.20* \\
HotpotQA & 68.80 & 72.24 & 72.72 & 70.90 & 71.70 & 70.55 & \textbf{74.01}* \\
MSMARCO & 42.00 & 43.56 & 44.77 &  \textbf{43.67} & 42.70 & 41.65 & 42.93* \\
NFCorpus & 36.00 & 37.73 & 39.34 & 38.77  & 38.50 & 38.16 & \textbf{39.74}* \\
NQ & 63.00 & 62.70 & 65.96 & 61.09 & 63.80 & 60.92 & \textbf{65.20}* \\
Quora & 86.00 & 88.34 & 87.65 & 86.84 & 87.30 & \textbf{87.95} & 87.65* \\
SCIDOCS & 16.10 & 19.66 & 19.45 & 19.26 & \textbf{20.80} & 20.02 & 19.52 \\
SciFact & 75.30 & 75.02 & 77.20 &  75.38 & \textbf{77.50} & 74.59 & 76.54 \\
TRECCOVID & 83.90 & 83.15 & 85.76 & 83.15 & 84.50 & 77.52 & \textbf{85.30}* \\
Touche2020 & 34.10 & 27.84 & 32.89 & 30.77 & 31.70 & 25.47 & \textbf{32.38} \\ \midrule
\textbf{Average} & 52.91 & 53.99 & 55.13 & 53.62 & \textbf{55.21} & 53.36 & 55.18* \\ \bottomrule
\end{tabular}
\label{table:beir_repllama_appdx}
\end{table*}
\begin{table*}[ht]
\small
\centering
\addtolength{\tabcolsep}{-3pt}
\caption{Performance (nDCG@10) on datasets from RAR-b when training decoder-only models. * indicates statistical significance over Promptriever $(p < 0.05)$ using the paired bootstrap test. For the Average score, we compute the overall p-value using Fisher's method.}
\begin{tabular}{lccccccc}
\toprule
\multirow{3}{*}{\textbf{Dataset}} & \textit{Llama2} & \multicolumn{2}{c}{\textit{Llama3}} & \multicolumn{4}{c}{\textit{Llama-3.1-Instruct}} \\ \cmidrule(l){2-2} \cmidrule(l){3-4} \cmidrule(l){5-8} 
 & RepLLaMA & RepLLaMA & RARe & \multicolumn{1}{c}{RepLLaMA} & Promptreiver & \multicolumn{2}{c}{RARe} \\
 Eval Format & $q^{\text{inst}}$ & $q^{\text{inst}}$ & $q^{\text{RARe}}$ & \multicolumn{1}{c}{$q^{\text{inst}}$} & $q^{\text{inst}}$ & $q^{\text{inst}}$ & $q^{\text{RARe}}$ \\ \midrule
 ARC-C & 11.79 & 11.65 & 13.48 & 11.68 & 14.63 & 13.24 & \textbf{15.02} \\
$\alpha$-NLI & 25.40 & 24.35 & 30.38 & 24.96 & 24.70 & 27.34 & \textbf{31.58}* \\
HellaSwag & 30.83 & 31.47 & 30.27 & 31.03 & \textbf{32.57} & 31.42 & 28.81 \\
  PIQA & 31.56 & 32.84 & 34.12 & 33.42 & 34.80 & 34.23 & \textbf{35.59}* \\
  Quail & 6.40 & 6.21 & 5.98 & 5.71 & \textbf{7.80} & 6.92 & 6.91\\
 SiQA & 2.82 & 2.61 & \textbf{3.87} & 2.75 & 3.53 & 2.18 & 3.14 \\
 TempReason-L1 & 1.49 & 1.75 & 3.61 & 2.05 & 4.32 & 4.84 & \textbf{6.59}* \\
WinoGrande& 51.58 & 37.11 & 57.01 & 42.01 & 45.25 & 44.72 & \textbf{61.69}* \\ \midrule
\textbf{Average} & 20.23 & 18.50 & 22.34 & 19.20 & 20.95 & 20.61 & \textbf{23.67}* \\ \bottomrule
\end{tabular}
\label{table:rarb_repllama_appdx}
\end{table*}
\subsection{Performance on BeIR and RAR-b}
\autoref{table:beir_appdx} and \autoref{table:rarb_appdx} provide detailed numbers on each dataset from BeIR and RAR-b respectively when training from retriever checkpoints. \autoref{table:beir_repllama_appdx} and \autoref{table:rarb_repllama_appdx} provide detailed numbers on each dataset from BeIR and RAR-b respectively when training from decoder-only LLMs.

\begin{table*}
\small
\centering
\addtolength{\tabcolsep}{-2pt}
\renewcommand{\arraystretch}{1.1}
\caption{Latency breakdown (in milliseconds per query) of each stage in the retrieval pipeline for $q^\text{inst}$ and $q^\text{RARe}$ evaluation settings. \textbf{\# Corpus} denote the number of documents and \textbf{Avg Q len.} denote the average number of query tokens split by whitespace. \autoref{table:eff_appdx} in the Appendix provides numbers on additional datasets.}

\begin{tabular}{lcccccccc}
\toprule
{\textbf{Dataset}} & \textbf{\# Corpus} & \textbf{Eval Setting} & \textbf{Avg Q len.} & \textbf{NN} & \textbf{Query} & \textbf{Search} & \textbf{Total} & \textbf{Inc.} \\ \midrule
\multirow{2}{*}{NFCorpus} & \multirow{2}{*}{3633} & $q^\text{inst}$ & 3.3 & 0 & {10.22} & {1.67} & {11.89} & - \\
& & $q^\text{RARe}$ & 866.0 & {0.62} & {473.65} & {1.76} & {476.04} & 40.04$\times$ \\ \midrule
\multirow{2}{*}{FiQA2018} & \multirow{2}{*}{57638} & $q^\text{inst}$ & 10.9 & 0 & 10.68 & 12.22 & 22.90 & - \\
& & $q^\text{RARe}$ & 1016.6 & {0.69} & {429.97} & {13.63} & {444.29} & 19.40$\times$ \\ \midrule
\multirow{2}{*}{TRECCOVID} & \multirow{2}{*}{171332} & $q^\text{inst}$ & 10.6 & 0 & {36.60} & {81.60} & {118.20} & - \\
& & $q^\text{RARe}$ & 722.54 & {6.20} & {435.60} & {86.00} & {527.80} & 4.47$\times$ \\ \midrule
\multirow{2}{*}{Touche2020} & \multirow{2}{*}{382545} & $q^\text{inst}$ & 6.6 & 0 & {28.98} & {189.59} & {218.57} & - \\
& & $q^\text{RARe}$ & 1287.8 & {4.08} & {822.86} & {214.29} & {1041.22} & 4.76$\times$ \\ \midrule
\multirow{2}{*}{Quora} & \multirow{2}{*}{522931} & $q^\text{inst}$ & 9.5 & 0 & {11.39} & {98.64} & {110.04} & - \\
& & $q^\text{RARe}$ & 129.5 & 0.32 & {53.03} & {98.26} & {151.61} & 1.38$\times$ \\ \midrule
\multirow{2}{*}{DBPedia} & \multirow{2}{*}{4635922} & $q^\text{inst}$ & 5.5 & 0 & {92.33} & {1470.95} & {1563.28} & - \\
& & $q^\text{RARe}$ & 158.2 & {0.48} & {115.53} & {1773.18} & {1889.18} & 1.21$\times$ \\ \midrule
\multirow{2}{*}{SciFact} & \multirow{2}{*}{5183} & $q^\text{inst}$ & 12.5 & 0 &  {15.07} &  {2.03} &  {17.10} & - \\
& & $q^\text{RARe}$ & 1250.7 &  {0.83} &  {707.83} &  {2.03} &  {710.70} & 41.56$\times$ \\ \midrule
\multirow{2}{*}{SCIDOCS} & \multirow{2}{*}{25657} & $q^\text{inst}$ & 9.4 & 0 &  {11.29} &  {5.74} &  {17.03} & - \\
& & $q^\text{RARe}$ & 901.1 &  {0.67} &  {354.82} &  {5.79} &  {361.28} & 21.21$\times$ \\ \midrule
\multirow{2}{*}{CQADupStack} & \multirow{2}{*}{38100} & $q^\text{inst}$ & 8.6 & 0 &  {9.13} &  {7.75} &  {16.88} & - \\
& & $q^\text{RARe}$ & 678.2 &  {1.15} &  {466.23} &  {7.79} &  {475.17} &  {28.15}$\times$ \\ \midrule
\multirow{2}{*}{ClimateFEVER} & \multirow{2}{*}{5416593} & $q^\text{inst}$ & 20.2 & 0 &  {16.98} &  {1124.36} &  {1141.34} & - \\
& & $q^\text{RARe}$ & 831.3  &  {2.31} &  {424.60} &  {1123.02} &  {1549.93} & 1.36$\times$ \\
\bottomrule
\end{tabular}
\label{table:eff_appdx}
\end{table*}

\subsection{Efficiency Evaluation}
\label{subsection:efficiency}

In Table \ref{table:eff_appdx}, we present a breakdown of the latency of each stage of the retrieval pipeline for both baseline ($q^{\text{inst}}$) and in-context ($q^{\text{inst+ic}}$) settings. We measure the total time required to obtain nearest-neighbour in-context examples (\textbf{NN}) from BM25, compute query embeddings (\textbf{Query}), and perform search with FAISS \citep{douze2024faisslibrary} with encoded query embeddings on the pre-computed document index (\textbf{Search}). We observe that the largest contributing factors to latency are the average length of input queries (\textbf{Avg Q len.}), and the size of the index (\textbf{\# Corpus}). The increased time for query encoding is not exclusive to our approach, since query expansion methods \cite{wang-etal-2023-query2doc, drozdov2023paradepassagerankingusing} also require encoding longer queries. \textbf{Moreover, the increased latency can be viewed as a form of test-time scaling, where the user trades-off throughput for performance depending on the preference. When no in-context examples are provided, we match the performance of the Instruct baseline}. For large query length and small corpus sizes, the in-context setting demonstrates a significant increase in total latency (19.40-40.04$\times$ for FiQA2018 and NFCorpus, respectively). However, for smaller average query lengths, this latency diminishes, as seen for Quora ($1.38\times$) and DBPedia ($1.21\times$). Moreover, the added latency due to the in-context setting also diminishes when the corpus size grows (which is the case in several real world settings), as the time required for search outweighs the  time to encode the query. For example, on Touche2020 with a larger corpus of 380K documents, the increase in latency is $4.76\times$ compared to FiQA2018 ($19.40\times$) for similar query lengths. Note that we did not perform any optimization that can further reduce latency. For instance, when the in-context examples are fixed, their KV representations can be pre-computed and cached, providing latency that is comparable to when only the original query is available. Furthermore, future research could explore several promising avenues, such as using efficient long-context retrievers \citep{saadfalcon2024benchmarkingbuildinglongcontextretrieval, zhang2024mambaretrieverutilizingmamba} as a backbone, or developing extractive and/or abstractive compression techniques of in-context examples.

\begin{table*}[ht]
\small
\centering
\caption{\textbf{Impact of the number of in-context examples ($k$) at inference time.} $k=5$ during training. All results are on \textit{E5-Mistral-Instruct}. In general, performance increases when increasing the number of examples, and the optimal number of examples can vary by task.}
\begin{tabular}{lcccccc}
\toprule
&&&& \# Examples \\ \cmidrule(lr){3-7} 
\textbf{Dataset}          & \textbf{Instruct (0)} & \textbf{0}    & \textbf{1}    & \textbf{3}    & \textbf{5}    & \textbf{10}   \\ \midrule
ArguAna        & 61.19           & \textbf{62.90} & 61.24          & 60.99         & 61.18         & 60.37         \\
ClimateFEVER    & \textbf{39.03}  & 38.99          & 38.27          & 37.97         & 37.50         & 37.67         \\
CQADupStack      & 44.82           & 45.57          & 47.49          & 48.33         & 48.46         & \textbf{48.48} \\
DBPedia          & 48.92           & 49.24          & 49.79          & 48.34         & 49.65         & \textbf{49.82} \\
FiQA2018         & 57.39           & 56.33          & \textbf{57.61} & 57.42         & 57.31         & 57.38         \\
NFCorpus         & 40.99           & 41.19          & 41.48          & 42.10         & 42.28         & \textbf{42.29} \\
SCIDOCS          & 17.94           & 18.71          & 19.83          & 20.17         & 20.19         & \textbf{20.20} \\
SciFact          & 77.28           & 77.11          & 83.56          & 84.45         & 84.79         & \textbf{85.12} \\
Touche2020       & 29.35           & 27.56          & 27.53          & 27.70         & 28.70         & \textbf{30.77} \\
TRECCOVID        & 72.89           & 77.03          & 76.96          & 78.99         & \textbf{79.58} & 78.77         \\ \midrule
\textbf{Average}          & 48.98           & 49.46          & 50.38          & 50.65         & 50.96         & \textbf{51.09} \\ \bottomrule
\end{tabular}
\label{table:eval_n_ic_examples_appdx}
\end{table*}

\begin{table*}[ht]
\small
\centering
\caption{\textbf{Impact of the number of in-context examples ($k$) during training and inference.} All results are on \textit{E5-Mistral-Instruct}. In general, performance increases when increasing the number of examples, and the optimal number of in-context examples can vary by task.}
\begin{tabular}{lcccccc}
\toprule
&&&& \# Examples \\ \cmidrule(lr){3-7} 
\textbf{Dataset}          & \textbf{Instruct (0)} & \textbf{0}    & \textbf{1}    & \textbf{3}    & \textbf{5}    & \textbf{10}   \\ \midrule
        Arguana    & 61.19 & \textbf{62.90} & 60.47 & 62.98 & 60.87 & 58.85 \\
        ClimateFEVER    & \textbf{39.03} & 38.99 & 37.94 & 36.45 & 37.50 & 36.54 \\
        CQADupStack & 44.82 & 45.57 & 46.76 & 47.12 & 48.46 & \textbf{48.92} \\
        DBPedia    & 48.92 & 49.24 & 47.70 & 49.05 & \textbf{49.65} & 47.95 \\
        FiQA2018   & \textbf{57.39} & 56.33 & 56.07 & 57.08 & 57.31 & 57.03 \\
        NFCorpus   & 40.99 & 41.19 & 40.67 & 40.77 & \textbf{42.28} & 42.24 \\
        SCIDOCS    & 17.94 & 18.71 & 20.01 & 19.28 & 20.19 & \textbf{21.54} \\
        SciFact    & 77.28 & 77.11 & 81.47 & 83.71 & 84.79 & \textbf{87.61} \\
        Touche2020 & \textbf{29.35} & 27.56 & 29.78 & 27.12 & 28.70 & 28.29 \\
        TRECCOVID  & 72.89 & 77.03 & 78.95 & 73.25 & 79.58 & \textbf{86.11} \\
        \midrule
        \textbf{Average} & 48.98 & 49.46 & 50.18 & 48.83 & 51.11 & \textbf{53.16} \\
        \bottomrule
    \end{tabular}
\label{table:n_ic_examples_appdx}
\end{table*}

\begin{table*}[ht]
\small
\centering
\addtolength{\tabcolsep}{-3.2pt}
\renewcommand{\arraystretch}{1.2}
\caption{\textbf{In-Context Format} Comparing variants of in-context example format on \textit{E5-Mistral-Instruct} during inference only. Training is done with the Regular format. Instruct refers to the baseline which does not use any in-context examples.}
\centering
\begin{tabular}{lcccccc}
\toprule
\multicolumn{1}{c}{} & \multicolumn{1}{c}{Instruct} & \multicolumn{5}{c}{\texttt{RARe}} \\ \cmidrule(lr){2-2} \cmidrule(lr){3-7}
\textbf{Dataset} & - & Query-Only & Doc-only & Shuffle-NC & Shuffle-C & Regular \\ \midrule
ArguAna & \textbf{61.19} & 57.36 & 60.35 & 55.64 & 60.49 & 60.87 \\
ClimateFEVER & 39.03 & 38.35 & 38.32 & 37.44 & \textbf{37.84} & 37.50 \\
CQADupStack & 44.82 & 39.56 & 48.43 & 47.70 & 48.27 & \textbf{48.46} \\
DBPedia & 48.92 & 49.14 & 49.69 & 49.72 & \textbf{50.04} & 49.65 \\
FiQA2018 & 57.39 & 55.67 & 56.85 & 56.64 & \textbf{57.41} & 57.31 \\
NFCorpus & 40.99 & 41.00 & 42.09 & 42.02 & 41.92 & \textbf{42.28} \\
SCIDOCS & 17.94 & 19.06 & 20.06 & 19.98 & \textbf{20.25} & 20.19 \\
SciFact & 77.28 & 77.46 & 81.88 & 81.51 & 82.20 & \textbf{84.79} \\
Touche2020 & \textbf{29.35} & 27.04 & 29.02 & 28.60 & 29.31 & 28.70 \\
TRECCOVID & 72.89 & 75.11 & 79.97 & 79.07 & \textbf{80.03} & 79.58 \\ \midrule
\textbf{Average} & 48.98 & 47.98 & 50.67 & 49.83 & 50.78 & \textbf{50.93} \\  \bottomrule
\end{tabular}
\label{table:eval_shuffle_appdx}
\end{table*}
\begin{table*}[ht]
\small
\centering
\addtolength{\tabcolsep}{-3.2pt}
\renewcommand{\arraystretch}{1.2}
\caption{\textbf{In-Context Format} Comparing variants of in-context example format on \textit{E5-Mistral-Instruct}. Instruct refers to the baseline which does not use any in-context examples.}
\centering
\begin{tabular}{lcccccc}
\toprule
\multicolumn{1}{c}{} & \multicolumn{1}{c}{Instruct} & \multicolumn{5}{c}{\texttt{RARe}} \\ \cmidrule(lr){2-2} \cmidrule(lr){3-7}
\textbf{Dataset} & - & Query-Only & Doc-Only & Shuffle-NC & Shuffle-C & Regular \\ \midrule
ArguAna & \textbf{61.19} & 58.88 & 57.54 & 60.17 & 58.97 & 60.87 \\
ClimateFEVER & \textbf{39.03} & 36.21 &35.59 & 30.83 & 35.71 & 37.50 \\
CQADupStack & 44.82 & 46.66 & 48.28 & 45.78 & 47.97 & \textbf{48.46} \\
DBPedia & 48.92 & 49.98 & 49.08 & \textbf{50.93} & 50.24 & 49.65 \\
FiQA2018 & \textbf{57.39} & 54.44 & 56.02 & 54.25 & 55.98 & 57.31 \\
NFCorpus & 40.99 & 41.42 & 41.62 & 41.17 & 41.78 & \textbf{42.28} \\
SCIDOCS & 17.94 & 20.04 & 20.12 & 20.35 & 20.11 & \textbf{20.19} \\
SciFact & 77.28 & 78.84 & 79.80 & 80.70 & 80.51 & \textbf{84.79} \\
Touche2020 & \textbf{29.35} & 28.09 & 29.01 & 29.18 & 28.97 & 28.70 \\
TRECCOVID & 72.89 & 79.54 & \textbf{83.29} & 82.14 & 82.97 & 79.58 \\ \midrule
\textbf{Average} & 48.98 & 49.41 & 50.04 & 49.55 & 50.32 & \textbf{50.93} \\  \bottomrule
\end{tabular}
\label{table:shuffle_appdx}
\end{table*}
\subsection{Choice of In-Context Examples}
\autoref{table:n_ic_examples_appdx} provides detailed numbers for varying in-context examples on all OOD BeIR tasks. \autoref{table:shuffle_appdx} provides detailed numbers for various prompt formats on all OOD BeIR tasks. 

\begin{table*}[t]
\small
\centering
\addtolength{\tabcolsep}{-3.8pt}
\renewcommand{\arraystretch}{1.2}
\caption{Performance (nDCG@10) on datasets from the BeIR benchmark \citealp{thakur2021beir} when training decoder-only model (\textit{Llama3}). Applying \texttt{RARe} with only in-context examples can lead to degradation of performance in the zero-shot setting ($q^\text{inst}$), but this is easily mitigated my including a mixture of $q^\text{inst}$ and $q^\text{RARe}$ data (30\% and 70\%) respectively.}
\begin{tabular}{lccccccccc}
\toprule
\textbf{Training} & \textbf{Eval} & \textbf{NQ} & \textbf{Quora} & \textbf{NFCorpus} & \textbf{SciFact} & \textbf{SCIDOCS} & \textbf{FiQA2018} & \multicolumn{1}{c}{\textbf{CQA}} & \textbf{Average} \\ \midrule
RepLLaMA-$q^\text{inst}$ & $q^\text{inst}$ & 62.70 & 88.34 & 37.73 & 75.02 & 19.66 & 44.31 & \multicolumn{1}{c}{42.59} & 52.91 \\ \midrule
& $q^\text{inst}$ & 39.64 & \textbf{88.39} & 35.42 &74.52 & \textbf{21.04} & 30.44 & 37.74 & 46.74 \\
\multirow{-2}{*}{\texttt{RARe}-$q^\text{RARe}$} & $q^\text{RARe}$& 65.19 & 86.79 & 38.87 & \textbf{78.41} & 19.70 & 46.58 & \textbf{43.75} & 54.18 \\ 
\rowcolor[HTML]{EFEFEF} & $q^\text{inst}$ & 63.68 & 87.84 & 38.06 & 76.07 &	20.11 & 46.02	& 42.99	& 53.54 \\
\rowcolor[HTML]{EFEFEF} \multirow{-2}{*}{\texttt{RARe}-$q^\text{inst}$ + $q^\text{RARe}$} & $q^\text{RARe}$& \textbf{65.96} & 87.65 & \textbf{39.34} & 77.20 & 19.45 & \textbf{47.13} & 42.96 & \textbf{54.24} \\ 
\bottomrule
\end{tabular}
\label{table:data_mix}
\end{table*}

\subsection{Mixture of Training Data}
In \autoref{table:data_mix}, we analyze the impact of training with only in-context examples when starting from decoder-only LLMs. As opposed to starting from existing retriever models, which have been trained without in-context examples, we observe that performance drops in the instruction-only setting. This can be largely mitigated by considering a mixture of in-context and instruction-only queries. 

\begin{figure*}
  \centering
  \begin{subfigure}{\linewidth}
    \centering
    \includegraphics[width=0.9\linewidth]{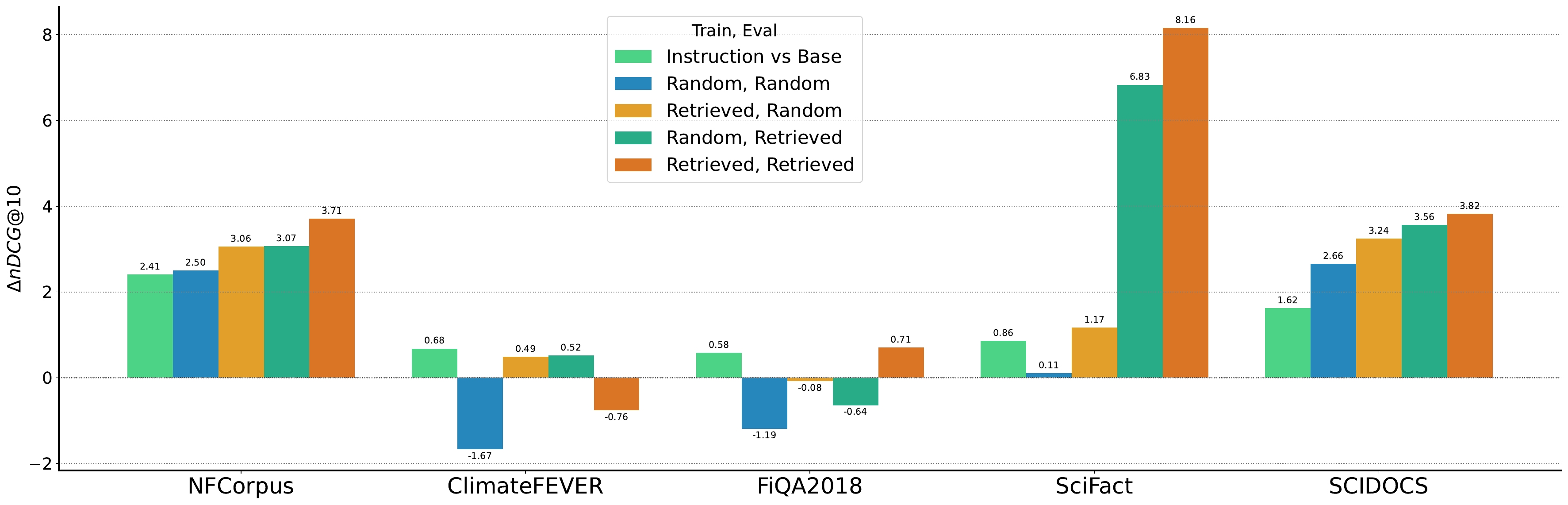}
    \label{fig:retvsrandom_a}
  \end{subfigure}
  \begin{subfigure}{\linewidth}
    \centering
    \includegraphics[width=0.9\linewidth]{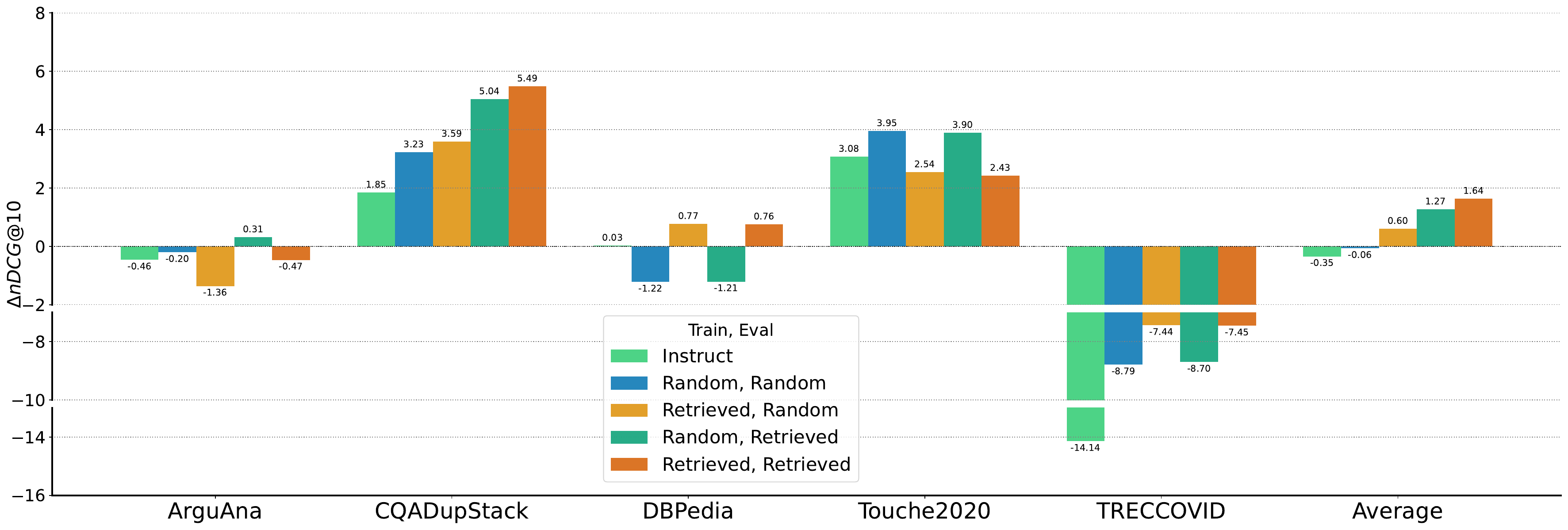}
    \label{fig:retvsrandom_b}
  \end{subfigure}
  \caption{\textbf{Retrieved vs. Random In-context Examples.} Change in performance ($\Delta$nDCG@10) on \textit{E5-Mistral-Instruct} with \texttt{RARe} ($q^\text{inst+ic}$) from the baseline setting ($q$ both during training and evaluation time). Using retrieved examples during training and/or inference enhance model performance in 7/10 datasets.}
  \label{fig:retvsrandom_appdx}
\end{figure*}

\begin{figure*}
  \centering
  \includegraphics[width=\linewidth]{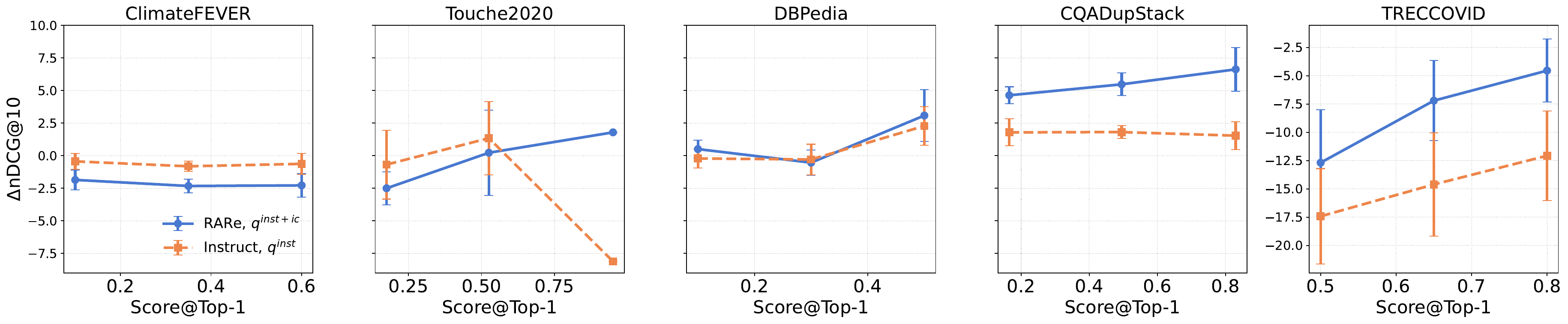}
  \caption{Change in performance ($\Delta$nDCG@10) from the base model (\textit{E5-Mistral-Instruct}) for varying similarity between the closest in-context example query and target query (Score@Top-1). Incorporation of semantically similar examples is beneficial on 3/5 datasets.}
  \label{fig:top_p_vs_ndcg_appdx}
\end{figure*}

\begin{table*}[t]
\small
\centering
\addtolength{\tabcolsep}{-3pt}
\caption{Performance (accuracy) with \textit{E5-Mistral-Instruct} on classification tasks, demonstrating the applicability of RARe on other embedding domains and tasks.\looseness=-1}
\begin{tabular}{lccccc}
\toprule
\textbf{Method} & \textbf{Banking77} & \textbf{Emotion} & \textbf{Intent} & \textbf{ToxicChat} & \textbf{Avg} \\ \midrule
Base            & 81.41              & 58.41            & 77.07                  & 81.67              & 74.64        \\
Instruct        & 83.27              & 58.00            & 77.57                  & 82.31              & 75.29        \\
\rowcolor[HTML]{EFEFEF} \texttt{RARe}            & \textbf{86.98}              & 57.18            & \textbf{78.11}                  & \textbf{83.99}              & \textbf{76.57}        \\ \bottomrule
\end{tabular}
\label{tab:clf}
\end{table*}
\subsection{Applicability to Other Embedding Tasks} In \autoref{tab:clf}, we report the performance on a subset of classification tasks from MTEB \cite{muennighoff-etal-2023-mteb}. We observe similar trends as retrieval, where providing in-context examples further enhances performance over providing task-specific instructions. The results underscore the applicability of \texttt{RARe} on embedding tasks beyond retrieval.

\begin{table}
\centering
\addtolength{\tabcolsep}{-4.3pt}
\begin{tabular}{lccccc}
\toprule
\textbf{Method} & \textbf{DBPedia} & \textbf{NFCorpus} & \textbf{SciFact} & \textbf{Covid} & \textbf{Touche2020} \\
\midrule
E5 \citep{wang2024textembeddings} & 40.7 & 35.0 & 70.4 & 74.1 & 30.9 \\
E5-Query2Doc \citep{wang-etal-2023-query2doc} & 42.4 & 35.2 & 67.5 & 75.1 & 31.7 \\
E5-Mistral-Instruct \citep{wang2024improving} & 48.92 & 40.99 & 77.28 & 72.89 & 29.35 \\
\rowcolor[HTML]{EFEFEF} \texttt{RARe} & 49.65 & 42.28 & 84.79 & 79.58 & 28.7 \\
\bottomrule
\end{tabular}
\caption{Comparisons against PRF method which use generative LLM to augment queries.}
\label{tab:prf_gen}
\end{table}


\begin{table}[t]
\small
\centering
\addtolength{\tabcolsep}{-4.7pt}
\caption{\textbf{Comparison against PRF methods.} The RARe pipeline i.e. BM25 based in-context selector and LLM-based retriever significantly outperforms methods where a generative LLM is used as a rewriter and BM25 as a retriever \citep{lei2024corpussteeredqueryexpansionlarge,jedidi2024zeroshotdenseretrievalembeddings}.}
\begin{tabular}{@{}lccccccccc@{}}
\toprule
\textbf{Method} & \textbf{Retriever} & \textbf{\begin{tabular}[c]{@{}c@{}}PRF\\ /ICL Model\end{tabular}} & \textbf{\begin{tabular}[c]{@{}c@{}}\#Used \\ Params\end{tabular}} & \textbf{Arguana} & \textbf{Covid} & \textbf{FiQA} & \textbf{SciFact} & \textbf{DBPedia} & \textbf{NFCorpus} \\ \midrule
ReDE-RF & BM25 & Mistral-7B & 7B & - & 65.60 & 29.30 & 66.90 & 37.60 & 35.50 \\
CSQE & BM25 & ChatGPT-3.5 & Unknown & 40.30 & 74.20 & 25.00 & 69.60 & 40.30 & - \\
\rowcolor[HTML]{EFEFEF} \texttt{RARe} & E5-Mistral-7B & BM25 & 7B & \textbf{60.87} & \textbf{79.58} & \textbf{57.31} & \textbf{84.79} & \textbf{49.65} & \textbf{42.28} \\ \bottomrule
\end{tabular}
\label{table:prf}
\end{table}
\begin{table}[t]
\centering
\caption{Performance of RARe against Instruct variant trained with $5\times$ steps. RARe performs better, especially on OOD settings, while Instruct begins to overfit on the in-doomain data.}
\begin{tabular}{lccc}
\toprule
\multirow{2}{*}{\textbf{Method}} & \multicolumn{2}{c}{\textbf{BeIR}} & \multirow{2}{*}{\textbf{RAR-b}} \\ \cmidrule(lr){2-3}
 & ID & OOD &  \\ \midrule
Instruct & 72.91 & 48.98 & 24.12 \\
Instruct (5x Steps) & 72.97 & 48.04 & 24.02 \\
\rowcolor[HTML]{EFEFEF} \texttt{RARe} (5 examples) & 72.98 & 50.93 & 25.79 \\ \bottomrule
\end{tabular}
\label{table:steps5}
\end{table}

  

\end{document}